\documentclass[12pt]{article} % For LaTeX2e
\usepackage{slashbox}

\usepackage{hyperref}
\usepackage{url}
\usepackage{booktabs}
\usepackage{graphicx}
\usepackage{subfig}
\usepackage{amsmath}
\usepackage{enumerate}
\usepackage{natbib}
\usepackage{booktabs} % For better-looking tables
\usepackage{caption} % For captions
\usepackage{array} % For centering cells
\usepackage{xspace}
\usepackage{authblk}

\usepackage{url} % not crucial - just used below for the URL 
\usepackage{soul}

\makeatletter
\let\ams@underbrace=\underbrace
\def\underbrace#1_#2{%
 \setbox0=\hbox{$\displaystyle#1$}%
 \ams@underbrace{#1}_{\parbox[t]{\the\wd0}{\centering #2}}%
}
\makeatother

\usepackage[T1]{fontenc}
\usepackage[utf8]{inputenc}
\usepackage{xcolor}	

\usepackage{microtype}

\usepackage{inconsolata}

\usepackage{times}
\usepackage{latexsym}
\usepackage{amssymb}
\usepackage{bbm}
\usepackage{bbold}
\usepackage{enumitem}

\usepackage[linesnumbered,algoruled,boxed,lined]{algorithm2e}
\usepackage{adjustbox}
\usepackage{stmaryrd}

\makeatletter 
\g@addto@macro{\@algocf@init}{\SetKwInOut{Parameter}{Parameters}} 
\makeatother
\SetKwComment{Comment}{//}{}
\RestyleAlgo{ruled}

\makeatletter
\def\@xfootnote[#1]{%
  \protected@xdef\@thefnmark{#1}%
  \@footnotemark\@footnotetext}
\makeatother

\DeclareMathOperator{\Cov}{Cov}

\title{From Narratives to Numbers:\\
Valid Inference Using Language Model Predictions\\
from Verbal Autopsy Narratives}

\author[1]{Shuxian Fan \thanks{denotes first authorship.}}
\author[2]{Adam Visokay \small{*}}
\author[1]{\\ \large{Kentaro Hoffman}}
\author[3]{Stephen Salerno}
\author[4]{\\ Li Liu}
\author[3]{Jeffrey T. Leek}
\author[1,2]{Tyler H. McCormick}

\affil[1]{\small Department of Statistics, 
University of Washington, 
Seattle, WA USA}
\affil[2]{\small Department of Sociology, 
University of Washington, 
Seattle, WA USA}
\affil[3]{\small Public Health Sciences Division, 
Fred Hutchinson Cancer Center, 
Seattle, WA USA}
\affil[4]{\small Bloomberg School of Public Health, 
Johns Hopkins University, 
Baltimore, MD USA}
\begin{document}

\maketitle

\begin{abstract}
In settings where most deaths occur outside the healthcare system, verbal autopsies (VAs) are a common tool to monitor trends in causes of death (COD). VAs are interviews with a surviving caregiver or relative that are used to predict the decedent's COD. Turning VAs into actionable insights for researchers and policymakers requires two steps (i) predicting likely COD using the VA interview and (ii) performing inference with predicted CODs (e.g. modeling the breakdown of causes by demographic factors using a sample of deaths). In this paper, we develop a method for valid inference using outcomes (in our case COD) predicted from free-form text using state-of-the-art NLP techniques. This method, which we call \texttt{multiPPI++}, extends recent work in ``prediction-powered inference'' to multinomial classification. We leverage a suite of NLP techniques for COD prediction and, through empirical analysis of VA data, we demonstrate the effectiveness of our approach in handling transportability issues. \texttt{multiPPI++} recovers ground truth estimates, regardless of which NLP model produced predictions and regardless of whether they were produced by a more accurate predictor like GPT-4-32k or a less accurate predictor like KNN. Our findings demonstrate the practical importance of inference correction for public health decision-making and suggests that if inference tasks are the end goal, having a small amount of contextually relevant, high quality labeled data is essential regardless of the NLP algorithm. 
\end{abstract}

\section{Introduction}

Verbal Autopsies (VAs) are a pivotal tool in public health research, particularly in contexts where access to formal medical records is limited \citep{soleman2006verbal, thomas2018verbal}. VA serves as a method to ascertain causes of death (COD) by collecting information from relatives or witnesses in circumstances necessitated by resource constraints, remote locations, or inadequate healthcare infrastructure~\citep{Baqui2006, Byass2012, Wahab2017, Blanco2021}.VA interviews include a structured yes/no questionnaire and an open text narrative where the interviewees provide additional information about the illness or death in their own words. 

After the interview, a COD is assigned either by clinician review or, more commonly, by using statistical algorithms. Finally, since financial and logistical constraints prevent VAs from being performed on all deaths, researchers and policymakers use statistical tools to summarize patterns in COD (e.g. the breakdown of infectious disease deaths by age, race, sex, etc).

\begin{figure}[!ht]
 \centering
 \textbf{\small The Workflow for Valid Inference Using \texttt{multiPPI++} for VA Narratives.}\par\medskip
 \includegraphics[width = 0.7\textwidth]{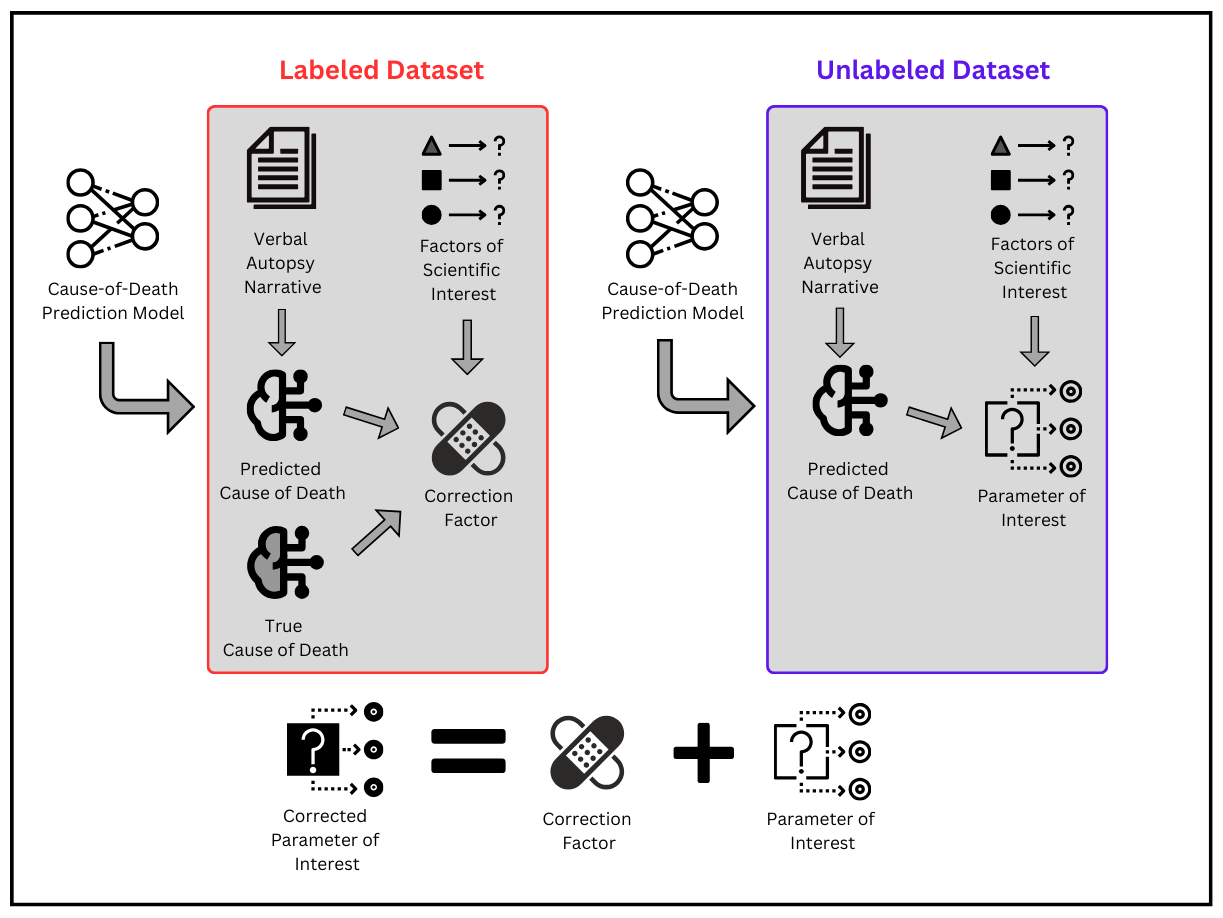}
 \caption{Overview of \texttt{multiPPI++} correction. Ground truth labels and predicted labels are used separately to perform the same inference task in Domain A. We use the difference between these estimates as a correction factor in Domain B where ground truth labels are not available.}
 \label{fig:overview}
\end{figure}

We propose a method for valid inference using the open text VA narratives, summarized in Figure \ref{fig:overview}. To do this, we address two pressing challenges. First, we use text narratives exclusively to classify COD. Structured VA interviews are long (typically around an hour) and emotionally taxing for respondents \citep{vainterview}. Narratives provide an opportunity for respondents to describe the circumstances around the death of someone they knew well in their own words, without needing to translate clinical jargon or answer irrelevant questions. Using narratives alone could also dramatically reduce the length of the interview, both minimizing impact on the respondent and allowing enumerators to collect more VAs. Recent work~\citep{danso2013linguistic, Blanco2021, manaka2022improving, cejudo2023cause} has begun exploring the potential of natural language processing (NLP) for COD classification in VA. Our work furthers this discussion by leveraging state-of-the-art NLP tools. Also, unlike prior NLP work which focuses on {\it predictive performance} \citep{naradowsky2012, NLPseptic2019, ye2021, peskoff-stewart-2023-credible}, our goal is to utilize these predictions in {\it statistical inference} to understand patterns in COD.

Second, we propose a method for inference that corrects for misclassification in CODs predicted by NLP. Training and evaluating COD prediction algorithms is extremely challenging \citep{Murray2011, Fottrell2010, Murray2014, clark2018quantifying, li2020using}, in part because building training sets with reliably labeled CODs is taxing. Clinicians could review VAs and assign causes, but this diverts critical resources from patient care and, since most deaths occur outside healthcare facilities, a physical autopsy is impossible. Hence, it is appealing to create a larger training set by combining the limited number of labeled VAs in one domain with other labeled VAs from different domains (e.g. locations). However, the diversity of cultural practices, linguistic nuances, and variations in interview techniques (not to mention potential biological factors) means that an algorithm trained on labeled VAs in one context may perform poorly in another. Our inferential methods must account for additional uncertainty that arises when most COD labels are predicted, not known. We also need to adjust for transportability so that VAs labeled in other contexts can be incorporated effectively into the training set. To do this, we extend prediction-powered inference \citep[\texttt{PPI++};][]{angelopoulos2023prediction, angelopoulos2023ppi} to settings of multinomial classification, showcasing its adaptability to a broader range of prediction problems. We propose an estimator, called \texttt{multiPPI++}, used to draw valid statistical inference using predictions from any arbitrary black-box machine learning model given access to a small subset of ground truth labels for multinomial data.

\section{Population Health Metrics Research Consortium Narratives} 
\label{sec:data}

The Population Health Metrics Research Consortium (PHMRC) dataset was collected in 2005 to provide a high quality, standardized dataset that includes ground truth COD labels from traditional autopsies, structured questionnaire responses, and free-text open narrative responses \citep{Murray2011}. It is one of a very small number of VA datasets with clinically validated causes. These data were curated from six sites across four countries: (\textit{Andhra Pradesh and Uttar Pradesh, India; Bohol, Philippines; Dar es Salaam and Pemba Island, Tanzania; and Mexico City, Mexico}). An example PHMRC narrative follows: \textit{``the deceased had been burnt and had lost mental balance and died within 1.5 hours of the accident''} (Ground Truth from Traditional Autopsy: \textbf{Fires}).

We focus our analysis on adult deaths, excluding those under 12 years of age, in the de-identified narratives released in \citet{Flaxman2018}. CODs are categorized into five broader groups: non-communicable diseases, communicable diseases, external causes, maternal causes, and AIDS or tuberculosis (AIDs-TB), based on International Classification of Diseases Tenth Revision (ICD-10) codes \citep{McCormick2016}. See Appendix \ref{sec:icd10} for the full mapping from ICD-10 codes to COD labels. 

\begin{figure}[!ht]
 \centering
 \textbf{ \small	Frequency Distribution of COD by Site}\par\medskip
 \includegraphics[width=\linewidth]{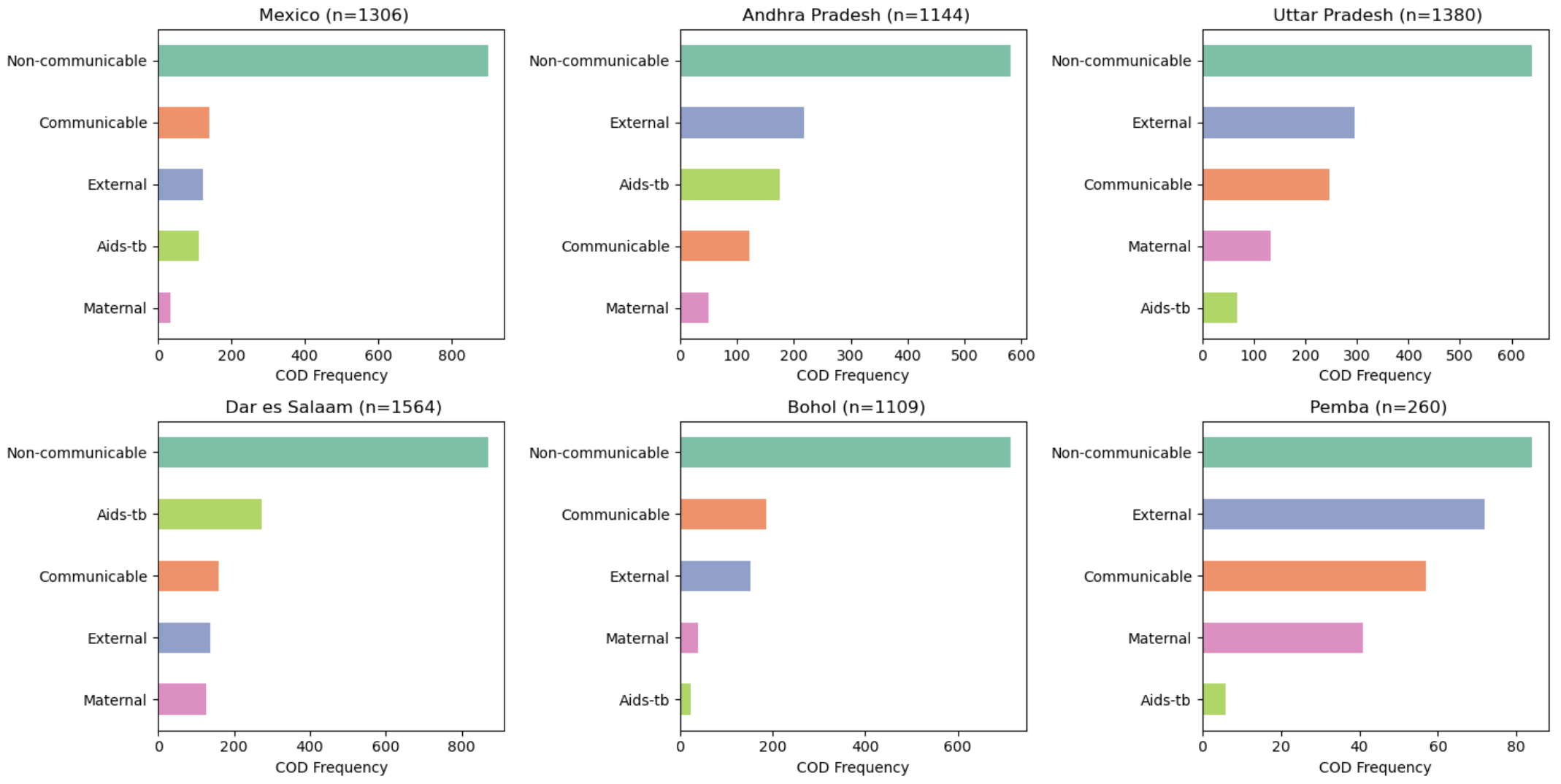}
 \caption{While non-communicable disease is the most common COD in each site, relative prevalence of each COD varies considerably. }
 \label{fig:cod_dist}
\end{figure}

Under these five broad classes, there still remains considerable heterogeneity in the COD distribution between sites. Figure \ref{fig:cod_dist} illustrates this phenomenon. While ``non-communicable'' diseases (e.g., various cancers, COPD, stroke) are the most common COD across all six sites, the second most common cause is site-specific. Namely, two sites identify communicable diseases (e.g., malaria, pneumonia) as the second most common COD, while external causes (e.g., suicide, homicide, traffic accidents) are reported in three sites and AIDs-TB in one site. In addition, the difference in rates between the first and second most common CODs varies considerably, with, for example, a 5:4 ratio in Pemba versus a 6:1 ratio in Mexico. 

These observations suggest that while information from one site may provide some insights into another, caution is warranted when extrapolating NLP model performance from one location to all others. In practice, this lack of transportability between sites poses a significant public health challenge and is well-documented in the VA literature on predicting COD \citep[e.g., see][]{McCormick2016}. However, without a means of addressing this issue for downstream statistical inference, it may be necessary to establish COD data collection systems at each new site of interest, an endeavor that would likely prove prohibitively expensive. We address this explicitly in Section \ref{sec:ppi}.

\section{Proposed Analytic Workflow}
\label{sec:methods}

\subsection{NLP for VA Narratives 
}
\label{sec:nlp}
The first aim of this work is to assess the predictive performance of contemporary NLP methods for classifying COD from free-text narratives, rather than structured interviews. To predict COD from the free-text narratives, we analyze a range of NLP techniques, including various methods based on a bag-of-words (BoW) representation and more sophisticated pre-trained models such as Bidirectional Encoder Representations from Transformers \citep[BERT;][]{bert_base} and large language models (LLMs) such as the recently popular Generative Pre-trained Transformer \citep[GPT model;][]{lai2023chatgpt}.

\begin{figure}[!ht]
 \centering
 \textbf{ \small	GPT-4 Zero-Shot Prompt Used for COD Prediction}\par\medskip
 \includegraphics[width = \textwidth]{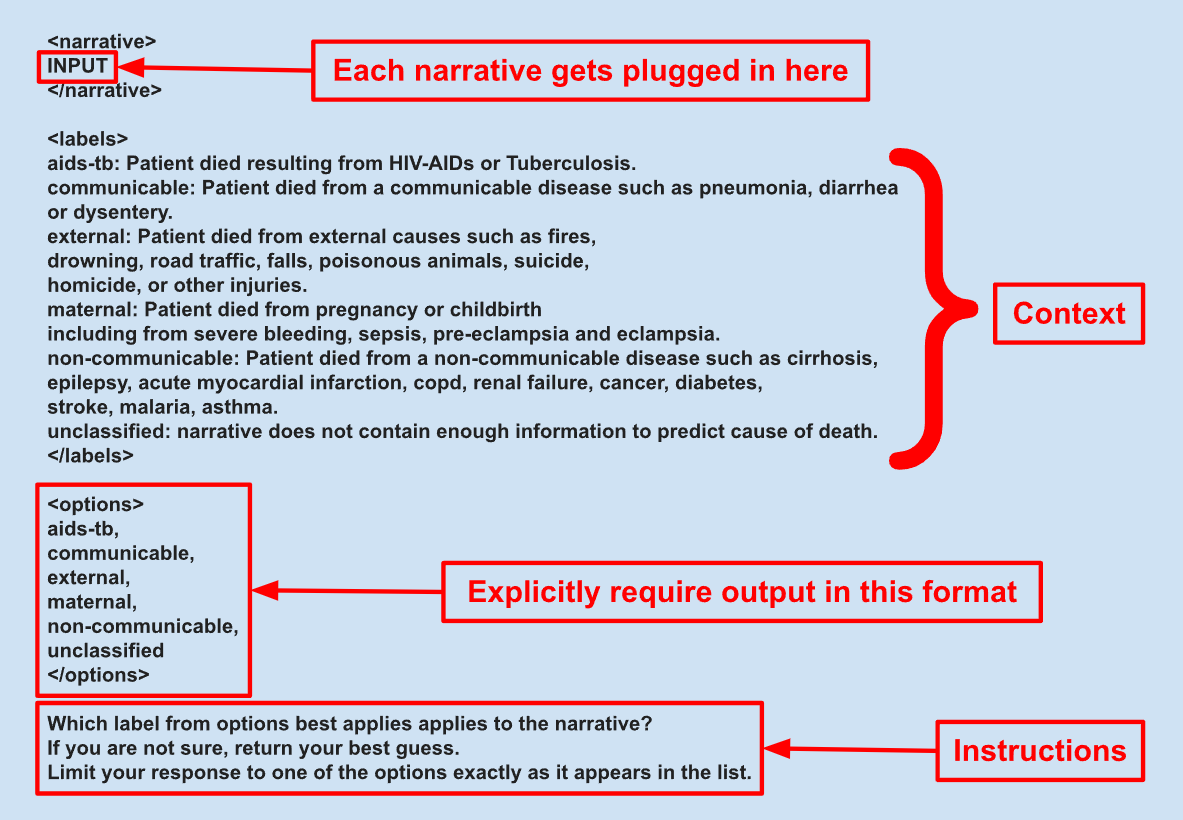}
 \caption{The zero-shot prompt explicitly tags the VA narrative, provides minimal context for each COD label, lists an explicit set of COD options to coerce a constrained output, and provides direct instructions pointing to the \texttt{<narrative><label><option>} tags.}
 \label{fig:prompt}
\end{figure}

To generate BoW representations of VA narratives, we employed \texttt{scikit-learn}'s \texttt{CountVectorizer}, \texttt{LabelEncoder}, and \texttt{TfidfVectorizer} \citep{kramer2016scikit,scikit-learn}, as well as \texttt{nltk}'s \texttt{word\_tokenize} \citep{millstein2020natural} for data pre-processing. After processing, the initial token count of $556,713$ was reduced to $218,804$. We then applied three supervised classifiers to predict COD using the BoW representations and ground truth cause of death labels: Support Vector Machine (SVM), K Nearest Neighbors (KNN), and Naive Bayes (NB). SVM was implemented with a third-degree linear kernel and $C=1.0$, while KNN used cosine distance and $9$ nearest neighbors, chosen by incrementing from $K=3$ until accuracy saturation. The $\text{BERT}_{BASE}$ encoder was configured with five transformer layers, to include two input layers of size 256, a main layer of size 768, an intermediate layer of size 512, and a fine-tuning output layer of size 5. As the VA narratives in our data were de-identified, we were able to further assess the predictive performance of OpenAI's GPT models \citep[GPT-3.5, GPT-3.5-turbo, GPT-4, and GPT-4-32K][]{biswas2023role}. After exploratory study, we selected GPT-4-32K for its larger context window. We used a zero-shot prompt (see Figure~\ref{fig:prompt}) with a temperature set to zero for all tasks. For all of the methods, we split the data into an 80\% training set and 20\% testing set, where we held out the true COD labels in the testing set to assess each method's predictive accuracy and F1-score. For $\text{BERT}_{BASE}$, training was conducted over two epochs. 

\subsection{Valid Statistical Inference with \texttt{multiPPI++} Correction } 
\label{sec:ppi} 
The second aim of this work addresses the challenge of conducting valid downstream inference on covariates associated with predicted COD labels. As long as the NLP model is not always 100\% accurate, uncritical inference using NLP predictions will always contain bias, which we must correct for before trusting the results. Note that while the PHMRC dataset is complete with ground truth COD labels from traditional, physical autopsies, this is often not practical due to resource constraints. Particularly in vulnerable areas without healthcare infrastructure, COD is abstracted solely from VA without access to means of verification. Recent works have shown that reifying algorithmically-derived outcomes (here, COD) for downstream inference can lead to biased coefficient estimates and anti-conservative inference \citep{wang2020methods, angelopoulos2023prediction, angelopoulos2023ppi, miao2023assumption, egami2023using, ogburn2021warning, hoffman2024we}.\\\\
To address this, we employ a recent technique known as prediction-powered inference \citep[\texttt{PPI};][]{angelopoulos2023prediction} and its extension, \texttt{PPI++}~\citep{angelopoulos2023ppi} and derive its appropriate form in the context of multiclass logistic regression. These methods operate by using a small subset of ground truth COD labels to {\it rectify} the coefficient estimates and standard errors from a regression based on predicted COD labels. To that end, assume that our data are comprised of $n$ {\it labeled} ($l$) observations, $\mathcal{L} = \{(Y_{l, i}, \hat{Y}^f_{l, i}, X_{l, i}, Z_{l, i}); i = 1, \ldots, n\}$, and $N$ {\it unlabeled} ($u$) observations, $\mathcal{U} = \{(\hat{Y}^f_{u, i}, X_{u, i}, Z_{u, i}); i = n + 1, \ldots, n + N\}$, where $Y$ denotes the true cause of death, $\hat{Y}^f$ denotes the cause of death predicted from an NLP function, $f(Z)$, $X$ denotes the covariates of interest, and $Z$ denotes the VA text narratives. In statistical inference, one can frame estimating statistical parameters, $\theta$, as minimizing an objective function of the form:

\begin{equation}
\theta^{\text{ \textit{Ground Truth}}} = \mathop{\arg \min}\limits_{\theta \in \mathbbm{R}^d} \mathbbm{E}[l_{\theta}(X, Y)],
\end{equation}
where $l_\theta: \mathcal{X}\times \mathcal{Y} \to \mathbbm{R}$ is a convex loss function in $\theta \in \mathbbm{R}^d$. In our case, the loss function $l_{\theta}(X, Y)$ is defined as the negative log conditional likelihood of the multinomial logistic regression. In practice, replacing the observed COD, $Y$, with $\hat{Y}^{f}$ yields what we refer to as a {\it Naive} estimate:

\begin{equation}
\theta^{Naive} = \mathop{\arg \min}\limits_{\theta \in \mathbbm{R}^d} \mathbbm{E}[l_{\theta}(X, \hat{Y}^{f})].
\end{equation}

The Naive estimate relies heavily on the assumption that the NLP model's predictions closely resemble the true CODs. 

\texttt{PPI++}, and our extension, {\tt multiPPI++}, seeks to utilize all available information from the unlabeled and labeled data, by minimizing the following objective function: 

\begin{equation}
\theta_{\lambda}^{PPI++} = \mathop{\arg \min}\limits_{\theta \in \mathbbm{R}^d} \left\{ \mathbbm{E}[l_\theta(X_{l}, Y_{l})] + 
\lambda \left(\mathbbm{E}[l_\theta(X_{u},\hat{Y}^f_{u})] - \mathbbm{E}[l_\theta(X_{l}, \hat{Y}_{l}^f)]\right) \right\},
\end{equation}
where $\lambda \in [0,1]$ is a tuning parameter which controls how much weight is placed on using the real labels versus the predicted labels. Note that this estimator combines the ground truth estimator from the labeled subset, which is statistically valid but may lack sufficient sample size, with the Naive estimator, which provides additional information and data from the unlabeled subset.

We extend the application of the \texttt{PPI++} approach from its logistic regression example to the multinomial case. ~\cite{angelopoulos2023ppi} mention this is possible, but do not provide details of the derivation, and their model has overparametrization issues. We instead derive a different parameterization that avoids identifiability issues, and we complete the derivation of \texttt{PPI++} correction for multiclass logistic regression. We refer to this variant as ``\texttt{multiPPI++}'', where, for the multinomial classification problem with $K$ classes and outcomes $Y_i\in \{0,\dots, K-1\}$, our loss function takes the form

\[
l_{\theta}(X,Y) = -\frac{1}{n}\sum_{i=1}^n X_i^T\theta_{Y_i} + \log\left(\sum_{k=1}^{K-1} e^{X_i^T\theta_{k}}\right).
\]

We present the \texttt{multiPPI++} adjustment procedures in Algorithm~\ref{alg:ppi_multi}. See Appendix~\ref{appendix:PPI} and~\ref{appendix:multiPPI} for complete derivations. 

\section{Inference with COD Predicted from VA Narratives }

\subsection{Experimental Setup}

\begin{figure}[!ht]
 \centering
 \textbf{\small ``Leave One Out'' Synthetic VA Transportability Experiment}\par\medskip
 \includegraphics[width=\linewidth]{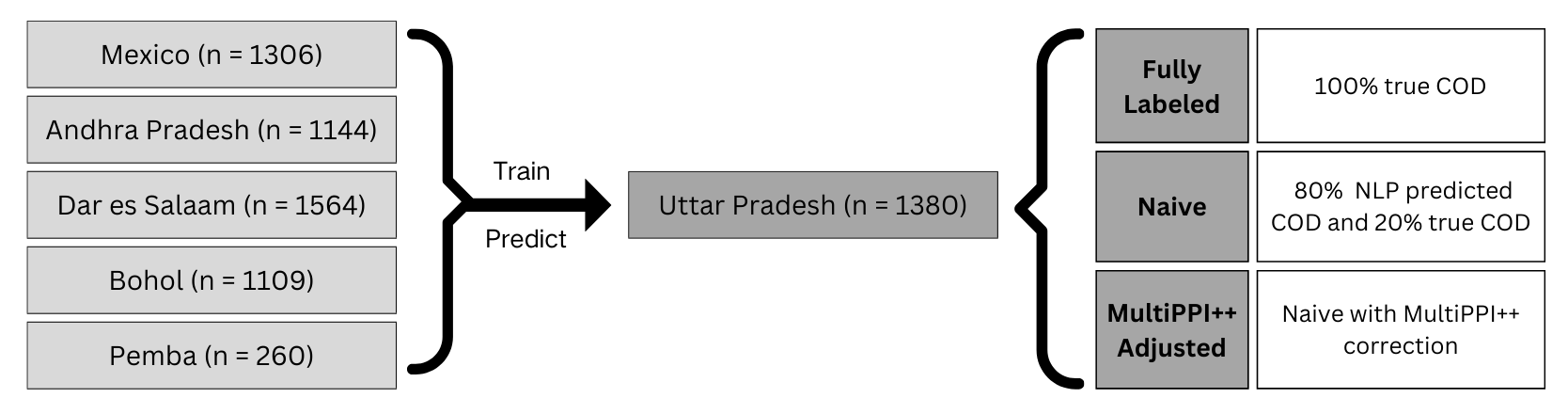}
 \caption{The classic and BERT models are trained with VA narratives from five other sites and used to predict COD from Uttar Pradesh narratives. The zero shot GPT-4 model predicts COD without any site specific narrative fine-tuning. Ground truth, Naive and \texttt{multiPPI++} corrected inference is performed using these Uttar Pradesh predictions.}
 \label{fig:exp-pic}
\end{figure}

To study the performance of the proposed methods, we designed an experiment in which we synthetically removed ground truth COD labels from the PHMRC dataset and replaced them with predicted COD from the NLP methods detailed above. Specifically, for each of the the six sites in the PHMRC dataset, we trained the NLP methods on the other five sites and used the resulting models to predict COD in the sixth site (see Figure~\ref{fig:exp-pic}). 

We compared the predicted outcomes to the withheld, true outcomes for the sixth site in terms of predicted accuracy and F1-score. We then carried forward these predictions to a downstream inferential model in which we regressed site-specific COD on the age of the decedent, to illustrate the performance of the {\tt multiPPI++} estimator. For each site, we retained 20\% of the labeled outcomes and replaced the remaining 80\% with the NLP predictions, to mirror the {\it labeled} and {\it unlabeled} subsets of the data necessary for the {\tt multiPPI++} approach. We compared the results of the proposed estimator to two additional estimators: (i) the `ground truth' estimator, which utilized the full set of true COD labels, and (ii) the `Naive' estimator, which used only the predicted COD, treating them as if they were observed. Note that, in many practical settings, the ground truth estimator is not feasible, but it is included here as a point of reference. Further, 

due to the variability in data between sites and the fact that the NLP model was not trained on this particular site, we anticipate that the regression coefficients estimated using the NLP-predicted COD will differ from those estimated using the true COD. 

We exemplify one representative set of experiments in which the Uttar Pradesh site is excluded from the NLP model training and used as the validation site to assess the NLP models' predictive performance and the downstream inferential results.

\subsection{NLP Prediction Results}

\begin{figure}[!ht]
 \centering
 \textbf{\small	Uttar Pradesh NLP Prediction Confusion Matrices}\par\medskip
 \subfloat[BoW with Na\"{i}ve Bayes]{%
  \includegraphics[width=.48\linewidth]{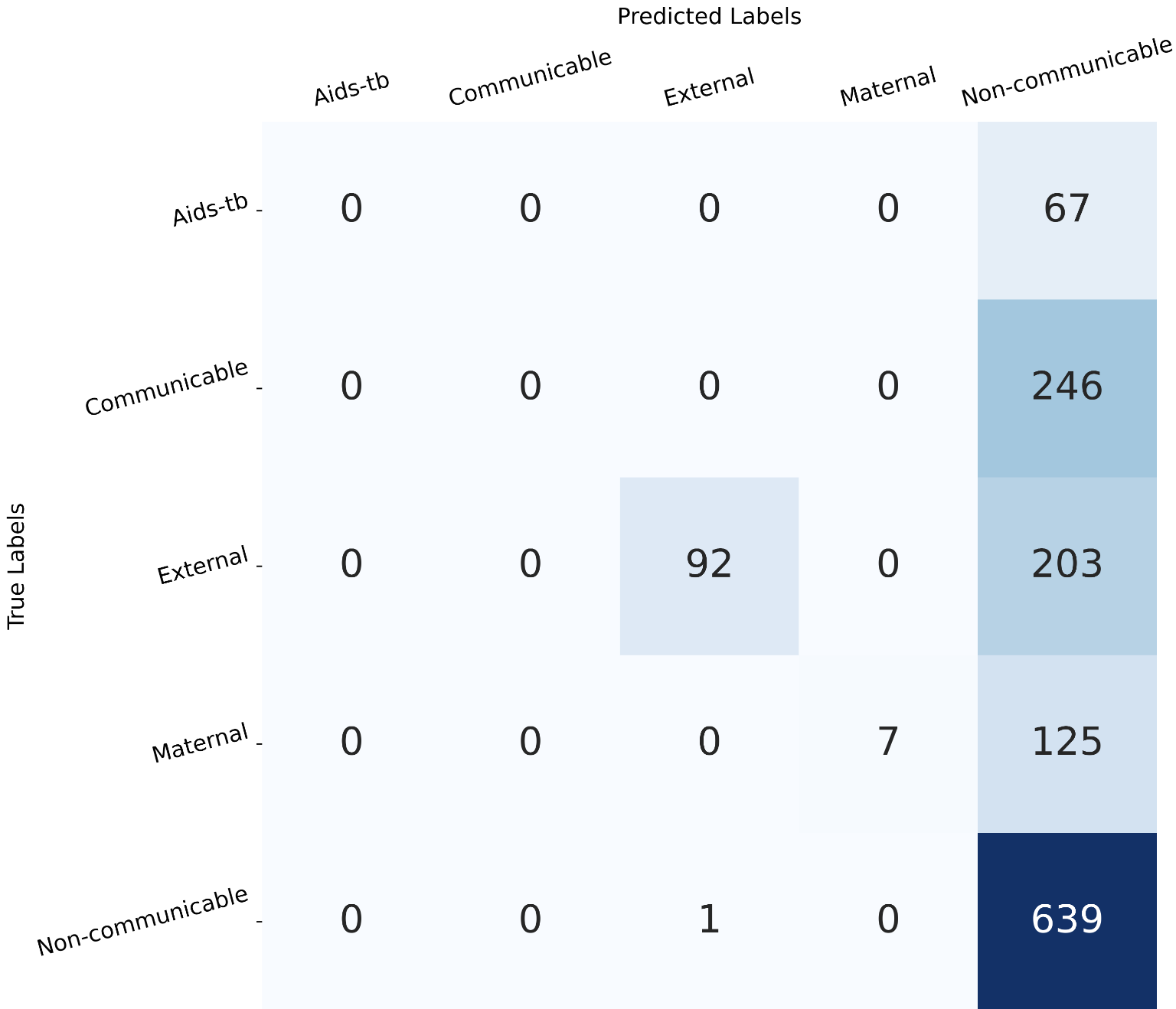}%
  \label{subfig:up_conf_matr-a}%
 }\hfill
 \subfloat[BoW with K Nearest Neighbors]{%
  \includegraphics[width=.48\linewidth]{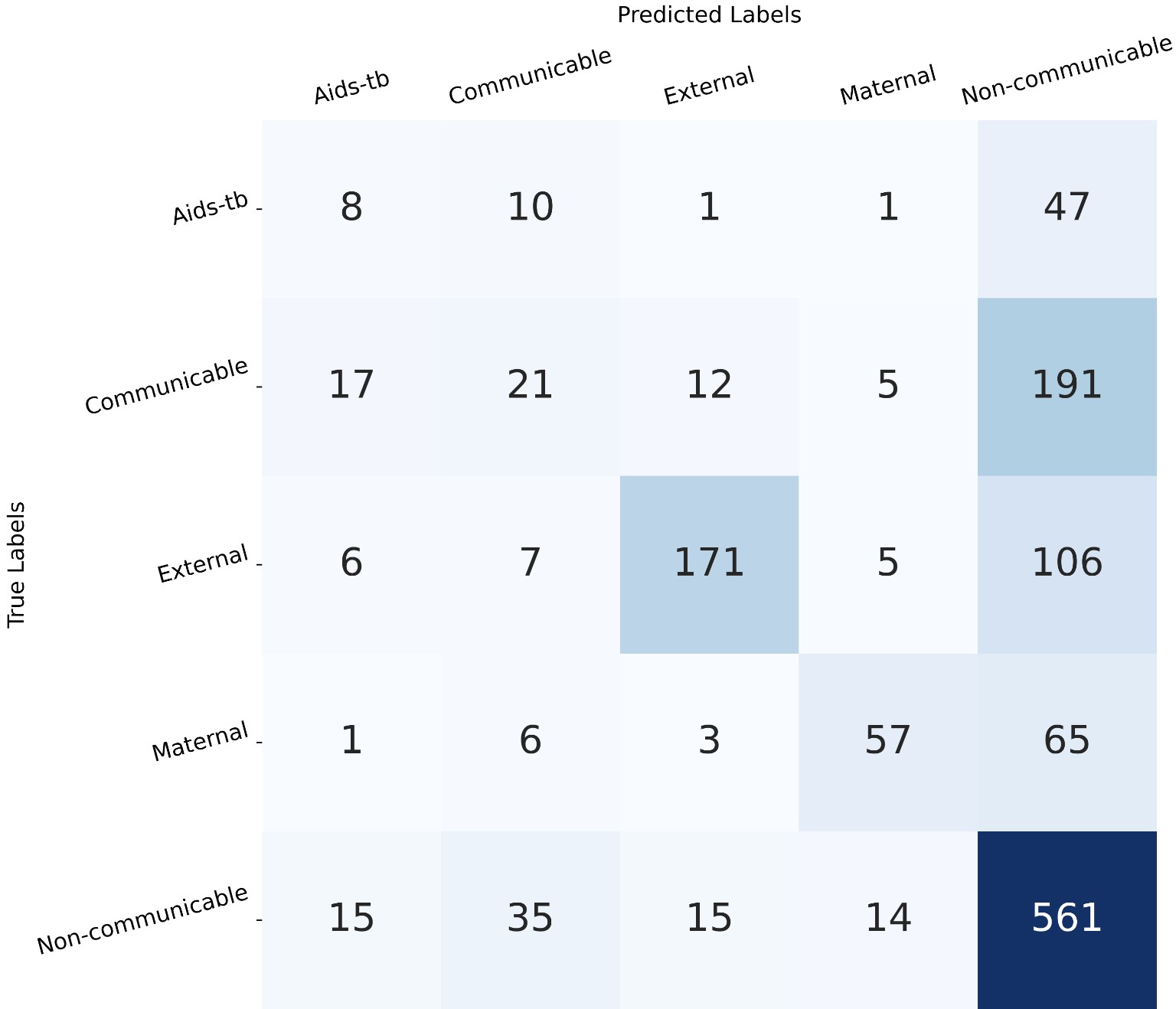}%
  \label{subfig:up_conf_matr-b}%
 }\\
 \subfloat[BoW with Support Vector Machine]{%
  \includegraphics[width=.48\linewidth]{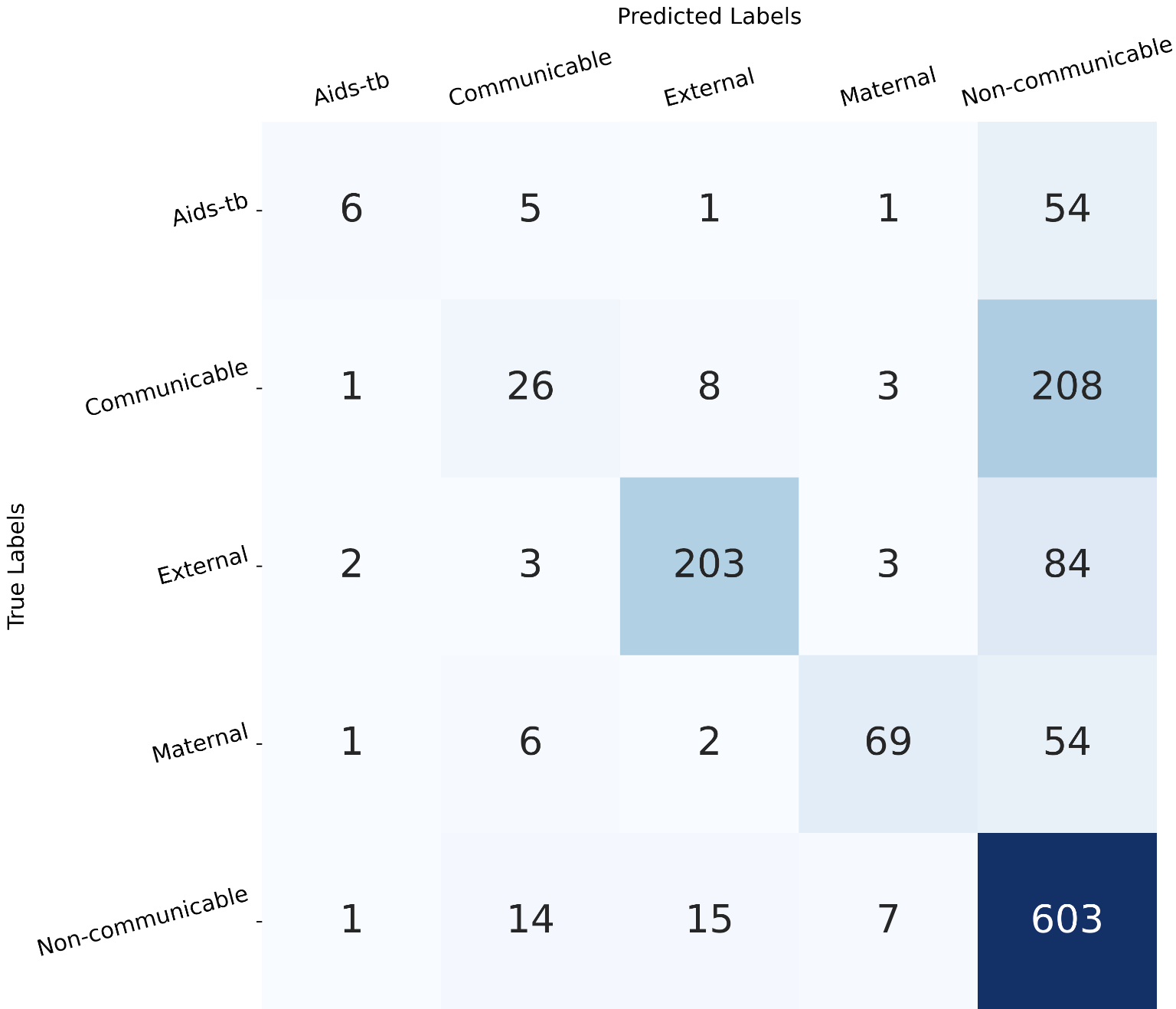}%
  \label{subfig:up_conf_matr-c}%
 }\hfill
 \subfloat[GPT-4]{%
  \includegraphics[width=.48\linewidth]{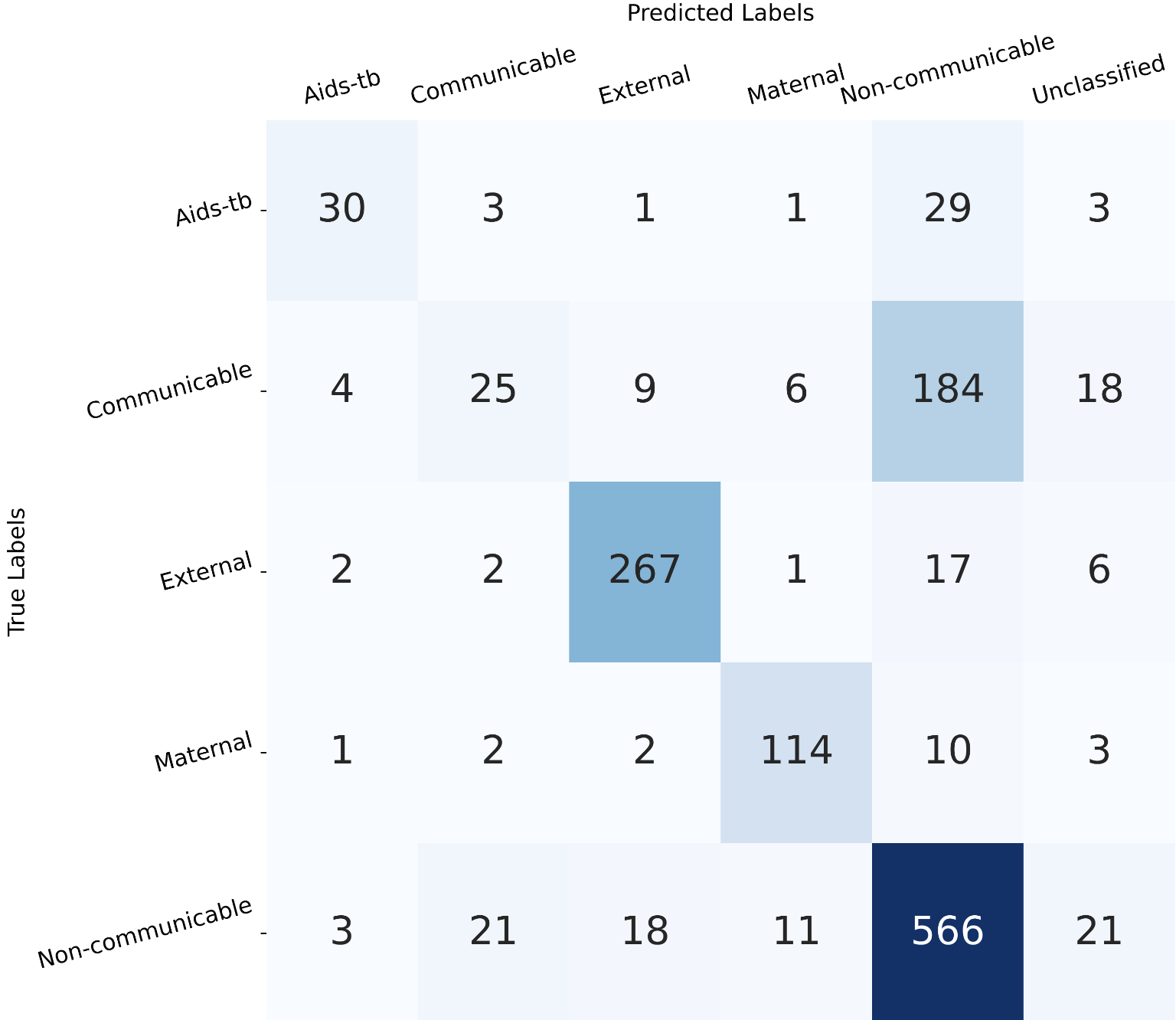}%
  \label{subfig:up_conf_matr-d}%
 }
 \caption{For VA narratives from Uttar Pradesh, most of the COD misclassifications are assigned non-communicable COD label. Naive Bayes mostly predicts non-communicable and achieves 0.60 accuracy in-part because non-communicable is overwhelmingly the most common ground truth COD.}
 \label{fig:up_conf_matr}
\end{figure}

First, using the ``leave one site out'' procedure described in Figure \ref{fig:exp-pic}, we trained each NLP model to predict COD from narratives on five sites, with the sixth site reserved as a validation set to use for assessing the methods. For the sake of brevity, we will focus our analysis one representative site, Uttar Pradesh. These results are summarized in Table~\ref{tab:f1-acc} and Figure~\ref{fig:up_conf_matr}, which contain the four models with highest accuracy. All models scored between 0.60-0.75 accurate, with GPT-4 yielding the highest accuracy (0.75). F1-scores were more uniform, with all but KNN getting F1-scores between 0.71-0.74. KNN scored 0.66. These results seem to point at GPT-4 either being slightly better or nearly equivalent to the other NLP methods we tested. GPT-4's predictions  matched the performance of the best examples from the VA literature, despite using only using only the VA text narratives and not the structured questionnaire \citep{james2011performance, Murray2014, Serina2015}. 

\begin{table}[!ht]
\centering
\renewcommand{\arraystretch}{1.5}
\begin{tabular}{|l|*{5}{l|}}\hline
\multicolumn{1}{|@{}l|}{\backslashbox[0pt][l]{Metrics}{ Models}}
&\makebox[5em]{BoW w/ SVM} & \makebox[5em]{BoW w/ KNN}& \makebox[5em]{BoW w/ NB} &\makebox[4em]{BERT} &\makebox[4em]{GPT-4}\\\hline\hline
Accuracy & 0.68 & 0.63 & 0.60 & 0.55 & 0.75 \\\hline
F1-Score & 0.74 & 0.66 & 0.72 & 0.71 & 0.73\\\hline
\end{tabular}
\caption{GPT-4 is far more accurate than competing NLP models and is among the best in terms of F1-score. Using only text narratives, GPT-4 performs on par with the best predictions in the VA literature, which are produced with much richer data in the form of structured questionnaires.}
\label{tab:f1-acc}
\end{table}

A unique feature of LLMs compared to other prediction algorithms is that they have the capacity, and in fact the tendency, to output class labels that do not exist in the ground labels. In our analysis, GPT-4 predicted 1503 out of 6763 COD labels as "unclassified". Upon manual review, these 1503 narratives contained no useful information for COD inference. The modal ``unclassified'' narrative was ``the interviewee thanked the interviewer'' (n=178). While the accuracies appear similar, GPT-4 correctly identified a flaw in the data cleaning in the PHMRC dataset, which suggest that the accuracies among the other methods may be overinflated. For each of the combination of six sites and five NLP model, we generated predictions and used this to generate regression coefficients.

\begin{figure}[!ht]
 \centering
 \textbf{\small	Inferential Experiment Results}\par\medskip
 \subfloat[BoW with Na\"{i}ve Bayes]{%
  \includegraphics[width=.5\linewidth]{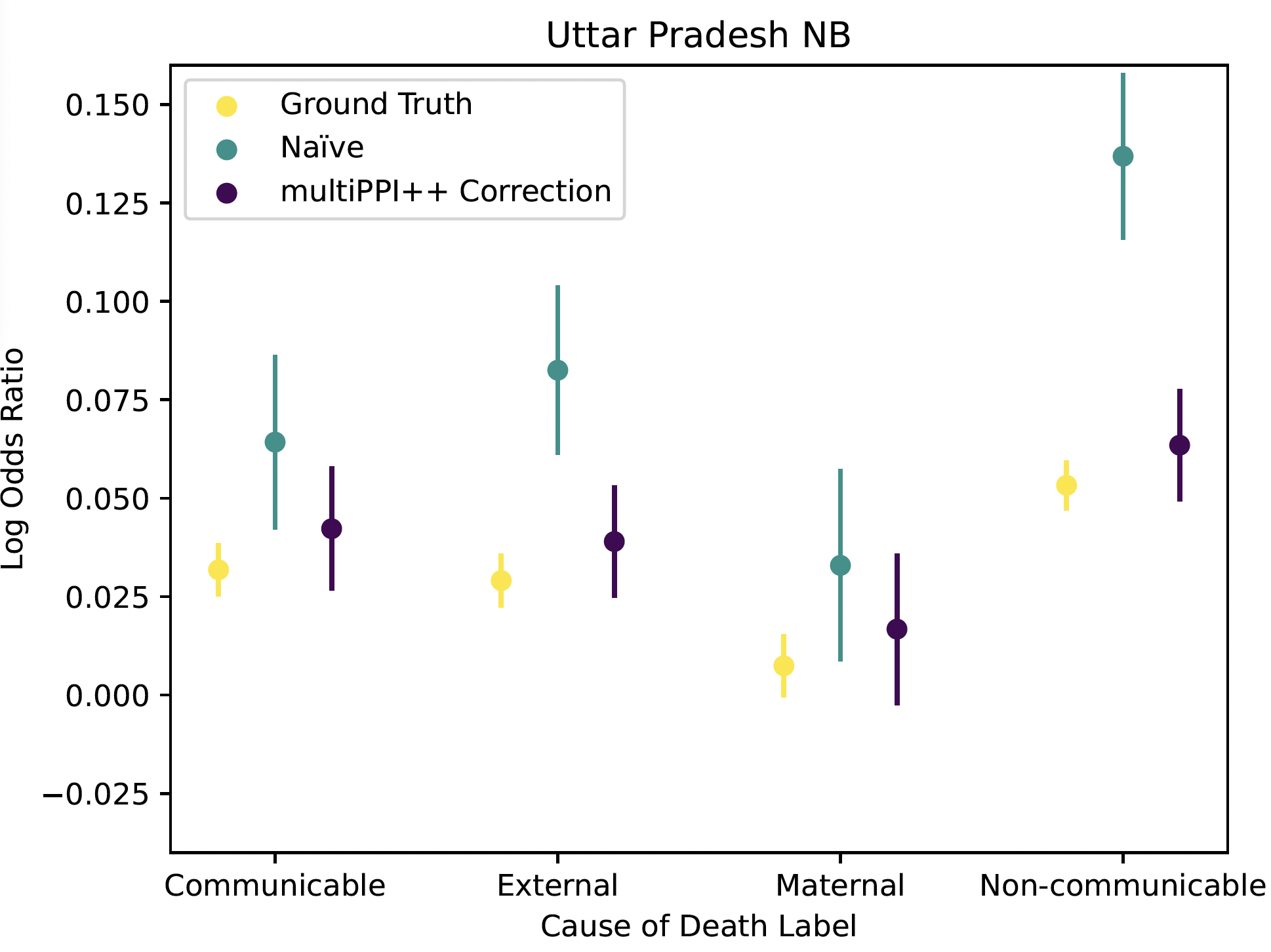}%
  \label{subfig:confint-a}%
 }\hfill
 \subfloat[BoW with K-Nearest-Neighbors]{%
  \includegraphics[width=.5\linewidth]{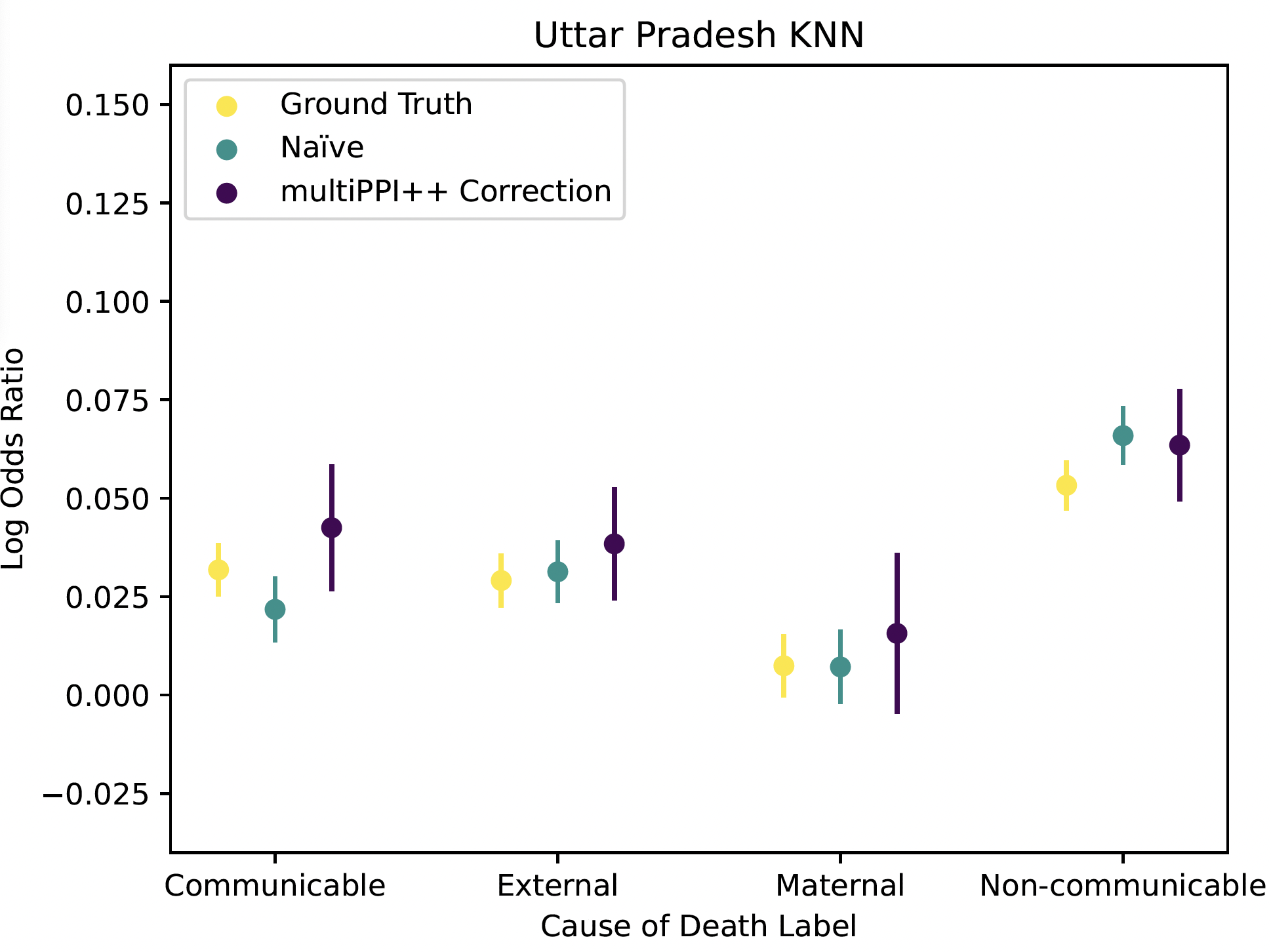}%
  \label{subfig:confint-b}%
 }\\
 \subfloat[BoW with Support Vector Machine]{%
  \includegraphics[width=.5\linewidth]{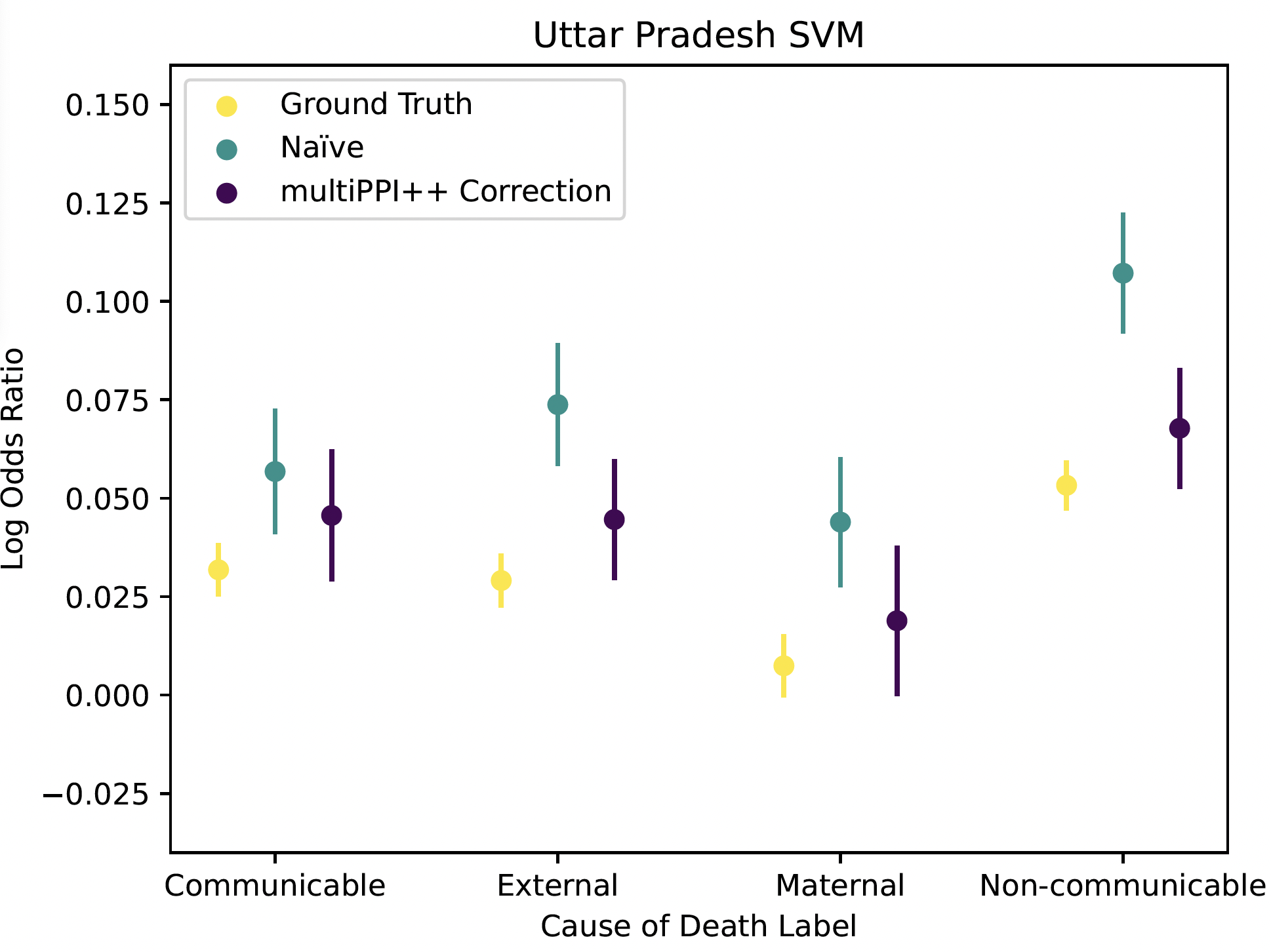}%
  \label{subfig:confint-c}%
 }\hfill
 \subfloat[GPT-4-32k]{%
  \includegraphics[width=.5\linewidth]{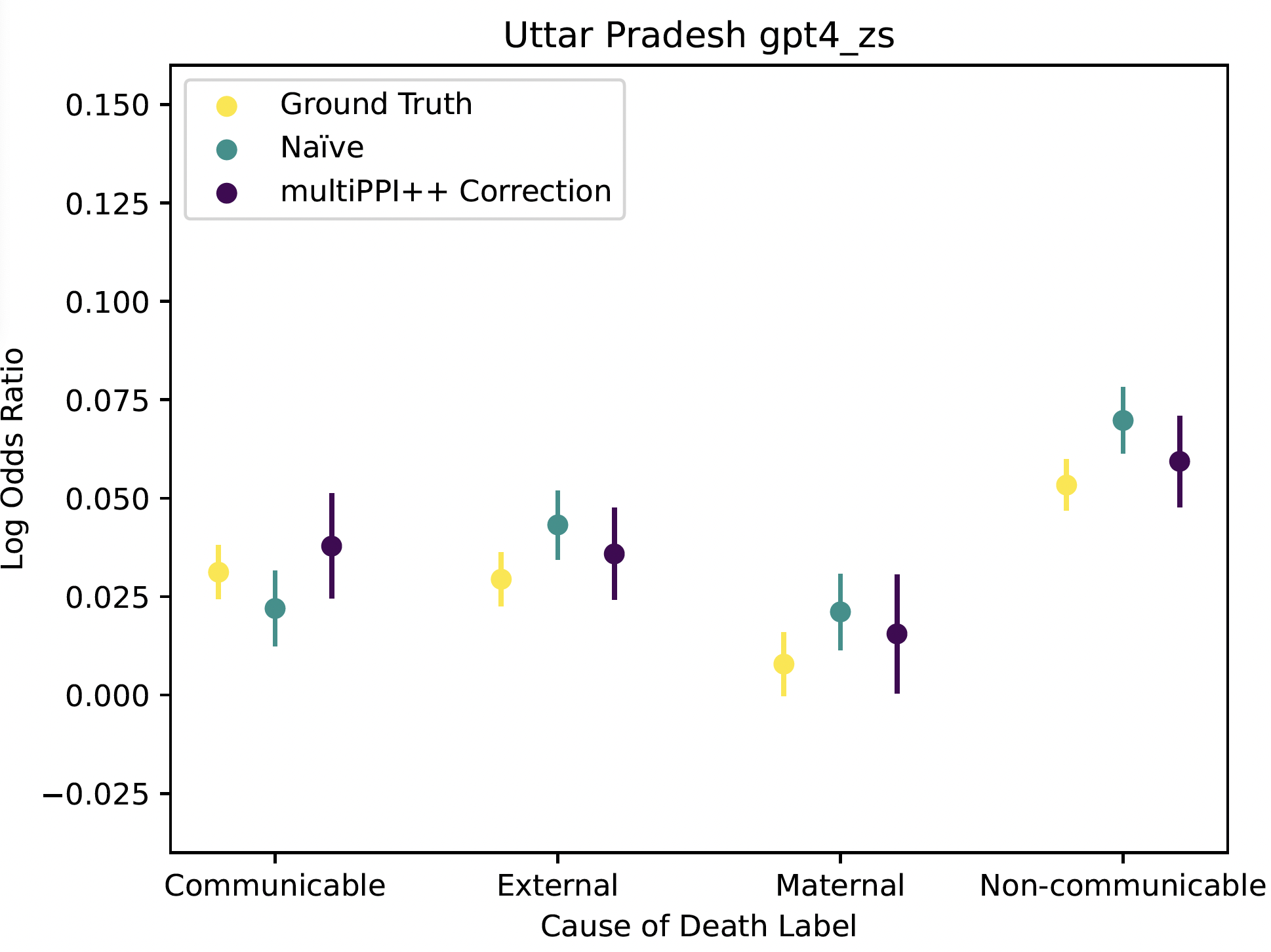}%
  \label{subfig:confint-d}%
 }
 \caption{Ground truth, Naive estimation, and \texttt{multiPPI++} correction of estimation of log odds ratios for multinomial regression for age in Uttar Pradesh. \texttt{multiPPI++} correction both recovers point estimates and reflects the increased uncertainty corresponding to the use of predicted COD rather than known COD with wider confidence intervals.}
 \label{fig:confint}
\end{figure}

For the sake of brevity, we present the four models with the best performance for the Uttar Pradesh site, NB, KNN, SVM, and GPT-4.  Figure \ref{fig:confint} displays point estimates and 95\% confidence intervals for the ground truth, Naive and \texttt{multiPPI++} corrected inference. 

\subsection{Inferential Model Results}

\begin{figure}[!ht]
 \centering
 \textbf{\small	No Apparent Association Between F1-scores and multiPP++ $\lambda$'s}\par\medskip
 \includegraphics[width=0.65\linewidth]{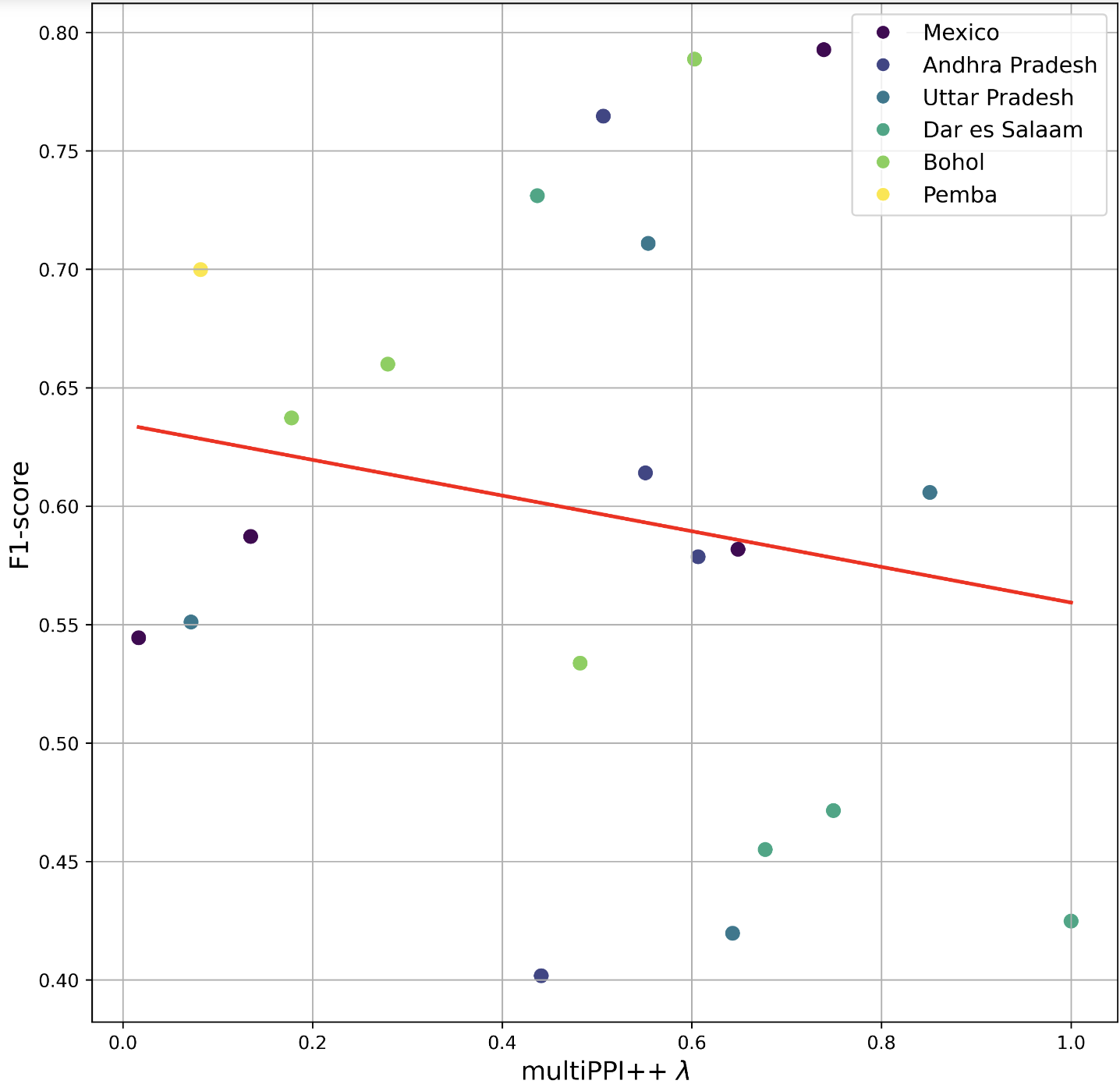}
 \caption{There appears no strong global relationship between \texttt{multiPPI++} $\lambda$ and F1-score, though associations do emerge when subsetting by site, hinting at a Yule-Simpson effect.}
 \label{fig:f1_lambda}
\end{figure}

In the case of NB and SVM, which had some of the lower F1-scores, there is a large discrepancy between the ground truth and the predicted values, demonstrating that replacing $80\%$ of the causes of death with predicted causes of death yields significantly different regression coefficients for the age of the decedent. \texttt{multiPPI++} correction rectifies estimates back to baseline, demonstrating its ability to account for (i) model inaccuracy and (ii) transportability bias. Interestingly, we observe that the Naive regression coefficients from KNN and GPT-4 are not as distant from the ground truth as the other two methods, even though KNN had the worst F1-score. Figure \ref{fig:f1_lambda} shows the association between F1-score and \texttt{multiPPI++} $\lambda$, which also fail to show any significant association. This implies that F1-score alone may not be a good predictor of the quality of downstream statistical inference. 

As for the widths of the confidence intervals, we observe that ground truth naturally had the thinnest, as it had the largest sample size and used real data points. On the other hand, the Naive estimate and \texttt{multiPPI++} correction show similar confidence interval widths. Since the widths of the confidence interval for \texttt{multiPPI++} is the sum of the widths of the Naive interval plus the width of the correction term, this seems to indicate that performing the \texttt{multiPPI++} correction does not seem to add a significant amount of uncertainty. This demonstrates that it can correct for transportability bias without much additional information cost. This is further evidenced by the seeming lack of association between the tuning parameter $\lambda$, which weights the use of the predicted outcomes in the downstream inferential model and the accuracy of the predictions. As $\lambda$ is a function of both the outcome and the associated features, we conclude that better predicted outcomes carry no additional useful information for parameter estimation.

\section{Discussion}

This paper uses a variety of NLP models to predict COD using VAs and, subsequently, develops a statistical method for valid inference using these predicted CODs. This research not only contributes to the advancement of predictive modeling in public health, but also underscores the significance of parameter interpretation in guiding decisions within diverse and dynamic health scenarios.     

Based on our research, there are several points that we believe are worthy of further investigation. First, we noticed that better predictive accuracy in the COD classifications did not necessarily translate to improved parameter estimation when used as a generative model for downstream statistical inference, at least not under the accuracy ranges examined in this paper. In situations where the user is ultimately interested in accurate parameter estimation, as is often the case in public health, care must be taken to differentiate between the most accurate model for prediction versus the most accurate model for parameter estimation, as it is possible that very cheap models such as KNN can be as useful as more expensive, newer models like GPT-4. For example, the cost to train and predict COD with BoW representations and KNN, NB and SVM was trivial, whereas performing the same task with GPT-4 cost \$3000USD. 

Second, the extreme class imbalance in COD drives many of our prediction results.  While it is not possible in practice to directly balance across CODs, there are broader design questions that could substantially improve inference. How, for example, should a researcher spend their very limited budget to label relevant VAs? And are there other ways of categorizing deaths that provide richer information than the five categories we use? Third, there is evidence that NLP techniques such as large language models perform worse in non-English languages \cite{Navigli2023, lai2023chatgpt, Hirschberg2015}, which is a major consideration since VAs are common in non-English settings. This could introduce bias to potential VA inference in other languages~\citep{vainterview}. This is a limitation of our simulation study, since it does not account for differential translation quality across languages. Future work might instead leverage NLP models specifically pre-trained for multiple languages or medical text \citep{peskoff2021, singhal_clinicalLLM2023}. 

\newpage

 \section*{Acknowledgements}
 We thank the University of Washington eScience institute for granting us Microsoft Azure computing credits used to produce all of our NLP predictions. We thank Aidan Andronicos, Robert Fatland, Musashi Hinck, Eddie Hock, Jesse Peirce, Denis Peskoff, Nels Schimek, Sander Schulhoff, and Brandon Stewart.

\newpage

% \bibliography{bibliography}

\begin{thebibliography}{40}
\providecommand{\natexlab}[1]{#1}
\providecommand{\url}[1]{\texttt{#1}}
\expandafter\ifx\csname urlstyle\endcsname\relax
  \providecommand{\doi}[1]{doi: #1}\else
  \providecommand{\doi}{doi: \begingroup \urlstyle{rm}\Url}\fi

\bibitem[Angelopoulos et~al.(2023a)Angelopoulos, Bates, Fannjiang, Jordan, and Zrnic]{angelopoulos2023prediction}
Anastasios~N. Angelopoulos, Stephen Bates, Clara Fannjiang, Michael~I. Jordan, and Tijana Zrnic.
\newblock Prediction-powered inference.
\newblock \emph{Science}, 382\penalty0 (6671):\penalty0 669–674, November 2023a.
\newblock ISSN 1095-9203.
\newblock \doi{10.1126/science.adi6000}.
\newblock URL \url{http://dx.doi.org/10.1126/science.adi6000}.

\bibitem[Angelopoulos et~al.(2023b)Angelopoulos, Duchi, and Zrnic]{angelopoulos2023ppi}
Anastasios~N. Angelopoulos, John~C. Duchi, and Tijana Zrnic.
\newblock Ppi++: Efficient prediction-powered inference, 2023b.
\newblock URL \url{https://arxiv.org/abs/2311.01453}.

\bibitem[Baqui et~al.(2006)Baqui, Darmstadt, Williams, Kumar, Kiran, Panwar, Srivastava, Ahuja, Black, and Santoshama]{Baqui2006}
A.~H. Baqui, G.~L. Darmstadt, E.~K. Williams, V.~Kumar, T.~U. Kiran, D.~Panwar, V.~K. Srivastava, R.~Ahuja, R.~E. Black, and M.~Santoshama.
\newblock {Rates, timing and causes of neonatal deaths in rural India: Implications for neonatal health programmes}.
\newblock \emph{Bulletin of the World Health Organization}, 84\penalty0 (9):\penalty0 706--713, 2006.
\newblock ISSN 00429686.
\newblock \doi{10.2471/BLT.05.026443}.

\bibitem[Biswas(2023)]{biswas2023role}
Som~S. Biswas.
\newblock Role of chat gpt in public health.
\newblock \emph{Annals of Biomedical Engineering}, 51\penalty0 (5):\penalty0 868–869, March 2023.
\newblock ISSN 1573-9686.
\newblock \doi{10.1007/s10439-023-03172-7}.
\newblock URL \url{http://dx.doi.org/10.1007/s10439-023-03172-7}.

\bibitem[Blanco et~al.(2021)Blanco, Perez, Casillas, and Cobos]{Blanco2021}
Alberto Blanco, Alicia Perez, Arantza Casillas, and Daniel Cobos.
\newblock {Extracting cause of death from verbal autopsy with deep learning interpretable methods}.
\newblock \emph{IEEE Journal of Biomedical and Health Informatics}, 25\penalty0 (4):\penalty0 1315--1325, apr 2021.
\newblock ISSN 21682208.
\newblock \doi{10.1109/JBHI.2020.3005769}.
\newblock URL \url{https://ieeexplore.ieee.org/abstract/document/9128025/}.

\bibitem[Byass et~al.(2012)Byass, Chandramohan, Clark, D'Ambruoso, Fottrell, Graham, Herbst, Hodgson, Hounton, Kahn, Krishnan, Leitao, Odhiambo, Sankoh, and Tollman]{Byass2012}
Peter Byass, Daniel Chandramohan, Samuel~J. Clark, Lucia D'Ambruoso, Edward Fottrell, Wendy~J. Graham, Abraham~J. Herbst, Abraham Hodgson, Sennen Hounton, Kathleen Kahn, Anand Krishnan, Jordana Leitao, Frank Odhiambo, Osman~A. Sankoh, and Stephen~M. Tollman.
\newblock {Strengthening standardised interpretation of verbal autopsy data: the new InterVA-4 tool.}
\newblock \emph{Global Health Action}, 5:\penalty0 1--8, 2012.
\newblock ISSN 16549880.
\newblock \doi{10.3402/GHA.V5I0.19281}.

\bibitem[Cejudo et~al.(2023)Cejudo, Casillas, P{\'e}rez, Oronoz, and Cobos]{cejudo2023cause}
Ander Cejudo, Arantza Casillas, Alicia P{\'e}rez, Maite Oronoz, and Daniel Cobos.
\newblock Cause of death estimation from verbal autopsies: Is the open response redundant or synergistic?
\newblock \emph{Artificial Intelligence in Medicine}, 143:\penalty0 102622, 2023.

\bibitem[Clark et~al.(2018)Clark, Li, and McCormick]{clark2018quantifying}
Samuel~J Clark, Zehang Li, and Tyler~H McCormick.
\newblock Quantifying the contributions of training data and algorithm logic to the performance of automated cause-assignment algorithms for verbal autopsy.
\newblock \emph{arXiv preprint arXiv:1803.07141}, 2018.

\bibitem[Danso et~al.(2013)Danso, Atwell, and Johnson]{danso2013linguistic}
Samuel Danso, Eric Atwell, and Owen Johnson.
\newblock Linguistic and statistically derived features for cause of death prediction from verbal autopsy text.
\newblock In \emph{Language Processing and Knowledge in the Web: 25th International Conference, GSCL 2013, Darmstadt, Germany, September 25-27, 2013. Proceedings}, pp.\  47--60. Springer, 2013.

\bibitem[Devlin et~al.(2018)Devlin, Chang, Lee, and Toutanova]{bert_base}
Jacob Devlin, Ming{-}Wei Chang, Kenton Lee, and Kristina Toutanova.
\newblock {BERT:} pre-training of deep bidirectional transformers for language understanding.
\newblock \emph{CoRR}, abs/1810.04805, 2018.
\newblock URL \url{http://arxiv.org/abs/1810.04805}.

\bibitem[Egami et~al.(2023)Egami, Jacobs-Harukawa, Stewart, and Wei]{egami2023using}
Naoki Egami, Musashi Jacobs-Harukawa, Brandon~M Stewart, and Hanying Wei.
\newblock Using large language model annotations for valid downstream statistical inference in social science: Design-based semi-supervised learning.
\newblock \emph{arXiv preprint arXiv:2306.04746}, 2023.

\bibitem[Flaxman et~al.(2018)Flaxman, Harman, Joseph, Brown, and Murray]{Flaxman2018}
Abraham~D. Flaxman, Lisa Harman, Jonathan Joseph, Jonathan Brown, and Christopher~J.L. Murray.
\newblock A de-identified database of 11, 979 verbal autopsy open-ended responses.
\newblock \emph{Gates Open Research}, 2:\penalty0 18, April 2018.
\newblock ISSN 2572-4754.
\newblock \doi{10.12688/gatesopenres.12812.1}.
\newblock URL \url{http://dx.doi.org/10.12688/gatesopenres.12812.1}.

\bibitem[Fottrell \& Byass(2010)Fottrell and Byass]{Fottrell2010}
E.~Fottrell and P.~Byass.
\newblock Verbal autopsy: Methods in transition.
\newblock \emph{Epidemiologic Reviews}, 32\penalty0 (1):\penalty0 38–55, March 2010.
\newblock ISSN 1478-6729.
\newblock \doi{10.1093/epirev/mxq003}.
\newblock URL \url{http://dx.doi.org/10.1093/epirev/mxq003}.

\bibitem[Hirschberg \& Manning(2015)Hirschberg and Manning]{Hirschberg2015}
Julia Hirschberg and Christopher~D. Manning.
\newblock Advances in natural language processing.
\newblock \emph{Science}, 349\penalty0 (6245):\penalty0 261–266, July 2015.
\newblock ISSN 1095-9203.
\newblock \doi{10.1126/science.aaa8685}.
\newblock URL \url{http://dx.doi.org/10.1126/science.aaa8685}.

\bibitem[Hoffman et~al.(2024)Hoffman, Salerno, Afiaz, Leek, and McCormick]{hoffman2024we}
Kentaro Hoffman, Stephen Salerno, Awan Afiaz, Jeffrey~T Leek, and Tyler~H McCormick.
\newblock Do we really even need data?
\newblock \emph{arXiv preprint arXiv:2401.08702}, 2024.

\bibitem[James et~al.(2011)James, Flaxman, Murray, and Research]{james2011performance}
Spencer~L James, Abraham~D Flaxman, Christopher~J Murray, and Consortium Population Health~Metrics Research.
\newblock Performance of the tariff method: validation of a simple additive algorithm for analysis of verbal autopsies.
\newblock \emph{Population Health Metrics}, 9\penalty0 (1):\penalty0 31, 2011.

\bibitem[Kramer \& Kramer(2016)Kramer and Kramer]{kramer2016scikit}
Oliver Kramer and Oliver Kramer.
\newblock Scikit-learn.
\newblock \emph{Machine learning for Evolution Strategies}, pp.\  45--53, 2016.

\bibitem[Lai et~al.(2023)Lai, Ngo, Veyseh, Man, Dernoncourt, Bui, and Nguyen]{lai2023chatgpt}
Viet~Dac Lai, Nghia~Trung Ngo, Amir Pouran~Ben Veyseh, Hieu Man, Franck Dernoncourt, Trung Bui, and Thien~Huu Nguyen.
\newblock Chatgpt beyond english: Towards a comprehensive evaluation of large language models in multilingual learning.
\newblock \emph{arXiv preprint arXiv:2304.05613}, 2023.

\bibitem[Li et~al.(2020)Li, McComick, and Clark]{li2020using}
Zehang~Richard Li, Tyler~H McComick, and Samuel~J Clark.
\newblock Using bayesian latent gaussian graphical models to infer symptom associations in verbal autopsies.
\newblock \emph{Bayesian Analysis}, 15\penalty0 (3):\penalty0 781, 2020.

\bibitem[Liu et~al.(2019)Liu, Greenstein, Sarma, and Winslow]{NLPseptic2019}
Ran Liu, Joseph~L. Greenstein, Sridevi~V. Sarma, and Raimond~L. Winslow.
\newblock Natural language processing of clinical notes for improved early prediction of septic shock in the icu.
\newblock In \emph{2019 41st Annual International Conference of the IEEE Engineering in Medicine and Biology Society (EMBC)}, pp.\  6103--6108, 2019.
\newblock \doi{10.1109/EMBC.2019.8857819}.

\bibitem[Manaka et~al.(2022)Manaka, van Zyl, and Kar]{manaka2022improving}
Thokozile Manaka, Terence van Zyl, and Deepak Kar.
\newblock Improving cause-of-death classification from verbal autopsy reports.
\newblock In \emph{Southern African Conference for Artificial Intelligence Research}, pp.\  46--59. Springer, 2022.

\bibitem[McCormick et~al.(2016)McCormick, Li, Calvert, Crampin, Kahn, and Clark]{McCormick2016}
Tyler~H. McCormick, Zehang~Richard Li, Clara Calvert, Amelia~C. Crampin, Kathleen Kahn, and Samuel~J. Clark.
\newblock Probabilistic cause-of-death assignment using verbal autopsies.
\newblock \emph{Journal of the American Statistical Association}, 111\penalty0 (515):\penalty0 1036–1049, July 2016.
\newblock ISSN 1537-274X.
\newblock \doi{10.1080/01621459.2016.1152191}.
\newblock URL \url{http://dx.doi.org/10.1080/01621459.2016.1152191}.

\bibitem[Miao et~al.(2023)Miao, Miao, Wu, Zhao, and Lu]{miao2023assumption}
Jiacheng Miao, Xinran Miao, Yixuan Wu, Jiwei Zhao, and Qiongshi Lu.
\newblock Assumption-lean and data-adaptive post-prediction inference.
\newblock \emph{arXiv preprint arXiv:2311.14220}, 2023.

\bibitem[Millstein(2020)]{millstein2020natural}
Frank Millstein.
\newblock \emph{Natural language processing with python: natural language processing using NLTK}.
\newblock Frank Millstein, 2020.

\bibitem[Murray et~al.(2011)Murray, Lopez, Shibuya, and Lozano]{Murray2011}
Christopher~JL Murray, Alan~D Lopez, Kenji Shibuya, and Rafael Lozano.
\newblock Verbal autopsy: advancing science, facilitating application.
\newblock \emph{Population Health Metrics}, 9\penalty0 (1), July 2011.
\newblock ISSN 1478-7954.
\newblock \doi{10.1186/1478-7954-9-18}.
\newblock URL \url{http://dx.doi.org/10.1186/1478-7954-9-18}.

\bibitem[Murray et~al.(2014)Murray, Lozano, Flaxman, Serina, Phillips, Stewart, James, Vahdatpour, Atkinson, Freeman, Ohno, Black, Ali, Baqui, Dandona, Dantzer, Darmstadt, Das, Dhingra, Dutta, Fawzi, Gómez, Hernández, Joshi, Kalter, Kumar, Kumar, Lucero, Mehta, Neal, Praveen, Premji, Ramírez-Villalobos, Remolador, Riley, Romero, Said, Sanvictores, Sazawal, Tallo, and Lopez]{Murray2014}
Christopher~JL Murray, Rafael Lozano, Abraham~D Flaxman, Peter Serina, David Phillips, Andrea Stewart, Spencer~L James, Alireza Vahdatpour, Charles Atkinson, Michael~K Freeman, Summer~Lockett Ohno, Robert Black, Said~Mohammed Ali, Abdullah~H Baqui, Lalit Dandona, Emily Dantzer, Gary~L Darmstadt, Vinita Das, Usha Dhingra, Arup Dutta, Wafaie Fawzi, Sara Gómez, Bernardo Hernández, Rohina Joshi, Henry~D Kalter, Aarti Kumar, Vishwajeet Kumar, Marilla Lucero, Saurabh Mehta, Bruce Neal, Devarsetty Praveen, Zul Premji, Dolores Ramírez-Villalobos, Hazel Remolador, Ian Riley, Minerva Romero, Mwanaidi Said, Diozele Sanvictores, Sunil Sazawal, Veronica Tallo, and Alan~D Lopez.
\newblock Using verbal autopsy to measure causes of death: the comparative performance of existing methods.
\newblock \emph{BMC Medicine}, 12\penalty0 (1), January 2014.
\newblock ISSN 1741-7015.
\newblock \doi{10.1186/1741-7015-12-5}.
\newblock URL \url{http://dx.doi.org/10.1186/1741-7015-12-5}.

\bibitem[Naradowsky et~al.(2012)Naradowsky, Riedel, and Smith]{naradowsky2012}
Jason Naradowsky, Sebastian Riedel, and David Smith.
\newblock Improving {NLP} through marginalization of hidden syntactic structure.
\newblock In Jun{'}ichi Tsujii, James Henderson, and Marius Pa{\c{s}}ca (eds.), \emph{Proceedings of the 2012 Joint Conference on Empirical Methods in Natural Language Processing and Computational Natural Language Learning}, pp.\  810--820, Jeju Island, Korea, July 2012. Association for Computational Linguistics.
\newblock URL \url{https://aclanthology.org/D12-1074}.

\bibitem[Navigli et~al.(2023)Navigli, Conia, and Ross]{Navigli2023}
Roberto Navigli, Simone Conia, and Bj\"{o}rn Ross.
\newblock Biases in large language models: Origins, inventory, and discussion.
\newblock \emph{Journal of Data and Information Quality}, 15\penalty0 (2):\penalty0 1–21, June 2023.
\newblock ISSN 1936-1963.
\newblock \doi{10.1145/3597307}.
\newblock URL \url{http://dx.doi.org/10.1145/3597307}.

\bibitem[Ogburn et~al.(2021)Ogburn, Rudolph, Morello-Frosch, Khan, and Casey]{ogburn2021warning}
Elizabeth~L Ogburn, Kara~E Rudolph, Rachel Morello-Frosch, Amber Khan, and Joan~A Casey.
\newblock A warning about using predicted values from regression models for epidemiologic inquiry.
\newblock \emph{American Journal of Epidemiology}, 190\penalty0 (6):\penalty0 1142--1147, 2021.

\bibitem[Pedregosa et~al.(2011)Pedregosa, Varoquaux, Gramfort, Michel, Thirion, Grisel, Blondel, Prettenhofer, Weiss, Dubourg, Vanderplas, Passos, Cournapeau, Brucher, Perrot, and Duchesnay]{scikit-learn}
F.~Pedregosa, G.~Varoquaux, A.~Gramfort, V.~Michel, B.~Thirion, O.~Grisel, M.~Blondel, P.~Prettenhofer, R.~Weiss, V.~Dubourg, J.~Vanderplas, A.~Passos, D.~Cournapeau, M.~Brucher, M.~Perrot, and E.~Duchesnay.
\newblock Scikit-learn: Machine learning in {P}ython.
\newblock \emph{Journal of Machine Learning Research}, 12:\penalty0 2825--2830, 2011.

\bibitem[Peskoff \& Stewart(2023)Peskoff and Stewart]{peskoff-stewart-2023-credible}
Denis Peskoff and Brandon Stewart.
\newblock Credible without credit: Domain experts assess generative language models.
\newblock In Anna Rogers, Jordan Boyd-Graber, and Naoaki Okazaki (eds.), \emph{Proceedings of the 61st Annual Meeting of the Association for Computational Linguistics (Volume 2: Short Papers)}, pp.\  427--438, Toronto, Canada, July 2023. Association for Computational Linguistics.
\newblock \doi{10.18653/v1/2023.acl-short.37}.
\newblock URL \url{https://aclanthology.org/2023.acl-short.37}.

\bibitem[Peskoff et~al.(2021)Peskoff, Hangya, Boyd-Graber, and Fraser]{peskoff2021}
Denis Peskoff, Viktor Hangya, Jordan Boyd-Graber, and Alexander Fraser.
\newblock Adapting entities across languages and cultures.
\newblock In Marie-Francine Moens, Xuanjing Huang, Lucia Specia, and Scott Wen-tau Yih (eds.), \emph{Findings of the Association for Computational Linguistics: EMNLP 2021}, pp.\  3725--3750, Punta Cana, Dominican Republic, November 2021. Association for Computational Linguistics.
\newblock \doi{10.18653/v1/2021.findings-emnlp.315}.
\newblock URL \url{https://aclanthology.org/2021.findings-emnlp.315}.

\bibitem[Serina et~al.(2015)Serina, Riley, Stewart, James, Flaxman, Lozano, Hernandez, Mooney, Luning, Black, Ahuja, Alam, Alam, Ali, Atkinson, Baqui, Chowdhury, Dandona, Dandona, Dantzer, Darmstadt, Das, Dhingra, Dutta, Fawzi, Freeman, Gomez, Gouda, Joshi, Kalter, Kumar, Kumar, Lucero, Maraga, Mehta, Neal, Ohno, Phillips, Pierce, Prasad, Praveen, Premji, Ramirez-Villalobos, Rarau, Remolador, Romero, Said, Sanvictores, Sazawal, Streatfield, Tallo, Vadhatpour, Vano, Murray, and Lopez]{Serina2015}
Peter Serina, Ian Riley, Andrea Stewart, Spencer~L. James, Abraham~D. Flaxman, Rafael Lozano, Bernardo Hernandez, Meghan~D. Mooney, Richard Luning, Robert Black, Ramesh Ahuja, Nurul Alam, Sayed~Saidul Alam, Said~Mohammed Ali, Charles Atkinson, Abdulla~H. Baqui, Hafizur~R. Chowdhury, Lalit Dandona, Rakhi Dandona, Emily Dantzer, Gary~L. Darmstadt, Vinita Das, Usha Dhingra, Arup Dutta, Wafaie Fawzi, Michael Freeman, Sara Gomez, Hebe~N. Gouda, Rohina Joshi, Henry~D. Kalter, Aarti Kumar, Vishwajeet Kumar, Marilla Lucero, Seri Maraga, Saurabh Mehta, Bruce Neal, Summer~Lockett Ohno, David Phillips, Kelsey Pierce, Rajendra Prasad, Devarsatee Praveen, Zul Premji, Dolores Ramirez-Villalobos, Patricia Rarau, Hazel Remolador, Minerva Romero, Mwanaidi Said, Diozele Sanvictores, Sunil Sazawal, Peter~K. Streatfield, Veronica Tallo, Alireza Vadhatpour, Miriam Vano, Christopher J.~L. Murray, and Alan~D. Lopez.
\newblock Improving performance of the tariff method for assigning causes of death to verbal autopsies.
\newblock \emph{BMC Medicine}, 13\penalty0 (1), December 2015.
\newblock ISSN 1741-7015.
\newblock \doi{10.1186/s12916-015-0527-9}.
\newblock URL \url{http://dx.doi.org/10.1186/s12916-015-0527-9}.

\bibitem[Singhal et~al.(2023)Singhal, Azizi, Tu, Mahdavi, Wei, Chung, Scales, Tanwani, Cole-Lewis, Pfohl, Payne, Seneviratne, Gamble, Kelly, Babiker, Sch\"{a}rli, Chowdhery, Mansfield, Demner-Fushman, Ag\"{u}era~y Arcas, Webster, Corrado, Matias, Chou, Gottweis, Tomasev, Liu, Rajkomar, Barral, Semturs, Karthikesalingam, and Natarajan]{singhal_clinicalLLM2023}
Karan Singhal, Shekoofeh Azizi, Tao Tu, S.~Sara Mahdavi, Jason Wei, Hyung~Won Chung, Nathan Scales, Ajay Tanwani, Heather Cole-Lewis, Stephen Pfohl, Perry Payne, Martin Seneviratne, Paul Gamble, Chris Kelly, Abubakr Babiker, Nathanael Sch\"{a}rli, Aakanksha Chowdhery, Philip Mansfield, Dina Demner-Fushman, Blaise Ag\"{u}era~y Arcas, Dale Webster, Greg~S. Corrado, Yossi Matias, Katherine Chou, Juraj Gottweis, Nenad Tomasev, Yun Liu, Alvin Rajkomar, Joelle Barral, Christopher Semturs, Alan Karthikesalingam, and Vivek Natarajan.
\newblock Large language models encode clinical knowledge.
\newblock \emph{Nature}, 620\penalty0 (7972):\penalty0 172–180, July 2023.
\newblock ISSN 1476-4687.
\newblock \doi{10.1038/s41586-023-06291-2}.
\newblock URL \url{http://dx.doi.org/10.1038/s41586-023-06291-2}.

\bibitem[Soleman et~al.(2006)Soleman, Chandramohan, and Shibuya]{soleman2006verbal}
Nadia Soleman, Daniel Chandramohan, and Kenji Shibuya.
\newblock Verbal autopsy: current practices and challenges.
\newblock \emph{Bulletin of the World Health Organization}, 84\penalty0 (3):\penalty0 239--245, 2006.

\bibitem[Surek-Clark(2020)]{vainterview}
Clarissa Surek-Clark.
\newblock {301Verbal Autopsy Interview Standardization Study: A report from the field}.
\newblock In \emph{{The Anthropological Demography of Health}}. Oxford University Press, 10 2020.
\newblock ISBN 9780198862437.
\newblock \doi{10.1093/oso/9780198862437.003.0011}.
\newblock URL \url{https://doi.org/10.1093/oso/9780198862437.003.0011}.

\bibitem[Thomas et~al.(2018)Thomas, D’Ambruoso, and Balabanova]{thomas2018verbal}
Lisa-Marie Thomas, Lucia D’Ambruoso, and Dina Balabanova.
\newblock Verbal autopsy in health policy and systems: a literature review.
\newblock \emph{BMJ global health}, 3\penalty0 (2):\penalty0 e000639, 2018.

\bibitem[Wahab et~al.(2017)Wahab, Choiriyyah, and Wilopo]{Wahab2017}
Abdul Wahab, Ifta Choiriyyah, and Siswanto~Agus Wilopo.
\newblock {Determining the cause of death: Mortality surveillance using verbal autopsy in Indonesia}.
\newblock \emph{American Journal of Tropical Medicine and Hygiene}, 97\penalty0 (5):\penalty0 1461--1468, 2017.
\newblock ISSN 00029637.
\newblock \doi{10.4269/AJTMH.16-0815}.

\bibitem[Wang et~al.(2020)Wang, McCormick, and Leek]{wang2020methods}
Siruo Wang, Tyler~H McCormick, and Jeffrey~T Leek.
\newblock Methods for correcting inference based on outcomes predicted by machine learning.
\newblock \emph{Proceedings of the National Academy of Sciences}, 117\penalty0 (48):\penalty0 30266--30275, 2020.

\bibitem[Ye et~al.(2021)Ye, Liu, Fu, and Neubig]{ye2021}
Zihuiwen Ye, Pengfei Liu, Jinlan Fu, and Graham Neubig.
\newblock Towards more fine-grained and reliable {NLP} performance prediction.
\newblock In Paola Merlo, Jorg Tiedemann, and Reut Tsarfaty (eds.), \emph{Proceedings of the 16th Conference of the European Chapter of the Association for Computational Linguistics: Main Volume}, pp.\  3703--3714, Online, April 2021. Association for Computational Linguistics.
\newblock \doi{10.18653/v1/2021.eacl-main.324}.
\newblock URL \url{https://aclanthology.org/2021.eacl-main.324}.

\end{thebibliography}

\bibliographystyle{bibstyle}

\vfill

\newpage

\appendix
\section{Appendix}

\subsection{ICD-10 COD Classification}
\label{sec:icd10}

\begin{center}
\begin{tabular}{ |p{6cm}||p{6cm}| }
 \hline
 \multicolumn{2}{|c|}{\textbf{Mapping 34 PHMRC All-Cause Mortality Labels to Five Broad COD Labels}} \\
 \hline
 \textbf{All-Cause Mortality Label} & \textbf{Broad COD Label} \\
 \hline
 cirrhosis & non-communicable \\
 epilepsy & non-communicable \\
 pneumonia & communicable \\
 copd & non-communicable \\
 acute myocardial infarction & non-communicable \\
 fires & external \\
 renal failure & non-communicable \\
 lung cancer & non-communicable \\
 maternal & maternal \\
 drowning & external \\
 other cardiovascular diseases & non-communicable \\
 aids & aids-tb \\
 other non-communicable diseases & non-communicable \\
 falls & external \\
 road traffic & external \\
 diabetes & non-communicable \\
 other infectious diseases & communicable \\
 tb & aids-tb \\
 suicide & external \\
 other injuries & external \\
 cervical cancer & non-communicable \\
 stroke & non-communicable \\
 malaria & non-communicable \\
 asthma & non-communicable \\
 colorectal cancer & non-communicable \\
 homicide & external \\
 diarrhea/dysentery & communicable \\
 breast cancer & non-communicable \\
 leukemia/lymphomas & non-communicable \\
 poisonings & external \\
 prostate cancer & non-communicable \\
 esophageal cancer & non-communicable \\
 stomach cancer & non-communicable \\
 bite of venomous animal & external \\
 \hline
\end{tabular}
\end{center}

\subsection{\texttt{PPI} and \texttt{PPI++}: overview}
\label{appendix:PPI}

We formalize our problem setting while reviewing the prediction-powered inference \texttt{PPI} and \texttt{PPI++} approach~\citep{angelopoulos2023prediction, angelopoulos2023ppi}. The goal of \texttt{PPI} is to estimate a $d$-dimensional parameter $\boldsymbol{\theta}$ with $n$ labeled data points $(X_{li}, Y_{li})\mathop{\sim}\limits^{\mathrm{iid}} \mathbbm{P}, i\in \{1,\dots, n\}$, and $N$ unlabeled data points $ X_{ui} \mathop{\sim}\limits^{\mathrm{iid}} \mathbbm{P}_{X}, i \in \{1,\dots, N\}$. 
% We assume that the feature distribution in both datasets is identical, that is $\mathbbm{P} = \mathbbm{P}_X \times \mathbbm{P}_{Y|X}$. 
Additionally, there is a black-box model $f$ that uses the features to predict the outcomes based on labeled and unlabeled data, denoted as $\hat{Y}_l^f = (\hat{Y}_{l1}^f, \dots, \hat{Y}_{ln}^f)$ and $\hat{Y}_{u}^f= (\hat{Y}_{u1}^f,\dots, \hat{X}_{uN}^f)$. The prediction rule $f$ is assumed to be independent of the observed data. While \texttt{PPI} performs inference on a correction term to incorporate these black-box predictions; \texttt{PPI++} improves the methodology to be more computationally efficient and develops a modification that adapts to the accuracy of the prediction rule $f$ by introducing an additional parameter $\lambda$ that interpolates between classical and prediction-powered inference. 

The starting point of \texttt{PPI} and \texttt{PPI++} approach is to define the population losses: \[ L(\theta) = \mathbbm{E}[l_{\theta}(X, Y)], \text{ and }\] \[L^f(\theta) = \mathbbm{E}[l_{\theta}(X, \hat{Y}^f)].\] It is recognized that \[ L^{f_{l}}(\theta) = \mathbbm{E}[l_{\theta}(X_l, \hat{Y}_l^f)] = \mathbbm{E}[l_{\theta}(X_u, \hat{Y}_u^f)] = L^{f_u}(\theta).\]

Therefore the ``rectified'' loss is defined as:
\[L^{PPI}(\theta) = L_n(\theta) + L_N^{f_u}(\theta)-L_n^{f_l}(\theta) \]
where
\[L_n(\theta) = \frac{1}{n}\sum_{i=1}^n l_{\theta}(X_{li}, Y_{li}), \] \[L_n^{f_l}(\theta) = \frac{1}{n}\sum_{i=1}^n l_{\theta}(X_{li}, \hat{Y}_{li}^f)), \text{ and }\]
\[L_N^{f_u}(\theta)= \frac{1}{N} \sum_{i=1}^N l_{\theta}(X_{ui}, \hat{Y}_{ui}^f).\]

The rectified loss leads to the prediction-powered point estimate:
\[\hat{\theta}^{PPI} = \mathop{\arg \min}\limits_{\theta \in \mathbbm{R}^d} L^{PPI}(\theta).\]

\texttt{PPI} constructs the confidence set $C_{\alpha}^{PPI} = \{\theta \in \mathbbm{R}^d \text{ s.t. } \boldsymbol{0}\in C_{\alpha}^{\nabla}(\theta)\}$ for true parameter $\theta^*$ by forming $(1-\alpha)$-confidence sets, $C_{\alpha}^{\nabla}(\theta)$, for the gradient of the loss $\nabla L^{PPI}(\theta)$. This approach works well for a fixed $\theta$ but suffers from computational issue when forming the confidence set for every $\theta \in \mathbbm{R}^d$. \texttt{PPI++} derives the asymptotic normality about $\hat\theta$. This result allows the construction of $(1-\alpha)$-confidence intervals $C_{\alpha}^{PPI} = \{\hat{\theta}_j^{PPI} \pm z_{1-\alpha/2}\times \hat{\sigma}_j/\sqrt{n}\}$, where $\hat{\sigma}_j^2$ is a consistent estimate of the asymptotic variance of $\hat{\theta}_j^{PPI}$ and $z_{q}$ is the $q$-quantile of standard normal distribution. Similar confidence intervals centered around $\hat{\theta}^{PPI}$ can be constructed when the inference is performed on the whole $\theta^*$ vector rather than a single coordinate $\theta_j^*$, which results in far more efficient algorithm. 

Moreover, \texttt{PPI++} generalizes \texttt{PPI} to allow the inference to adapt to the accuracy of the supplied predictions, yielding estimations that are never worse than the classical inference. This approach is called ``power tuning'' by incorporating a tuning parameter $\lambda\in [0,1]$ in the rectified loss, leading to the following estimator:
\[\hat{\theta}_{\lambda}^{\texttt{PPI++}} = \mathop{\arg \min}\limits_{\theta} L_{\lambda}^{\texttt{PPI++}}(\theta),\]
where \[L_{\lambda}^{\texttt{PPI++}}(\theta) =L_n(\theta) + 
\lambda (L_N^{f_u}(\theta)-L_n^{f_l}(\theta)).\]

\texttt{PPI++} recovers the original \texttt{PPI} approach when $\lambda = 1$ and reduces to classical inference when $\lambda = 0$. One can adaptively choose a data-dependent tuning parameter $\hat{\lambda}$ to maximize the estimation and inferential efficiency. 

\begin{algorithm}[!ht]
\KwIn{labeled $K$-category COD data $\{(X_{li}, Y_{li})\}_{i=1}^n$, unlabeled data $\{X_{ui}\}_{i=1}^N$, NLP model $f$, significance level $\alpha\in (0,1)$, coefficient index $j\in[d(K-1)]$} 
\begin{enumerate}
 \item Optimally select tuning parameter $\hat\lambda$ \quad \Comment{\ set tuning parameter} 
 \item $\hat{\theta}_{\hat{\lambda}}^{\texttt{m}} = \mathop{\arg \min}\limits_{\theta \in \mathbbm{R}^{d(K-1)}} L_{\hat{\lambda}}^{\texttt{m}}(\theta)$ \quad \Comment{\ \texttt{multiPPI++} estimator }
 \item $\hat{H} = \frac{1}{N+n}(\sum_{i=1}^n \psi''(X_{li}^T\hat{\theta}_{\hat{\lambda}}^{\texttt{m}})X_{li}X_{li}^T + \sum_{i=1}^N\psi''(X_{ui}^T\hat{\theta}_{\hat{\lambda}}^{\texttt{m}})X_{ui}X_{ui}^T)$, where $\psi(\theta, x) = \log\left(\sum_{k=1}^{K-1} e^{x^T\theta_{k}}\right), \theta_k\in{\mathbbm{R}^d}$ \quad \Comment{\ empirical Hessian}
 \item $\hat{\Sigma} = \hat{H}^{-1}(\frac{n}{N}\hat{V}_{f} + \hat{V}_{\Delta})\hat{H}^{-1}$, where \item[] $\hat{V}_{f} = \hat{\lambda}^2 \widehat{\Cov}_{N+n}((\psi'(X_{li}^T\hat{\theta}_{\hat{\lambda}}^{\texttt{m}})-\hat{Y}_{li}^f))X_{li})$ and
 $\hat{V}_{\Delta} = \widehat{\Cov}_{n}((1-\hat{\lambda})(\psi'(X_{li}^T\hat{\theta}_{\hat{\lambda}}^{\texttt{m}}) + (\hat{\lambda}\hat{Y}_{li}^f - Y_{li})X_{li})$ \quad \Comment{\ covariance estimator} 
 
\end{enumerate}
\KwOut{\par Prediction-powered point estimates $\hat{\theta}_{\hat{\lambda}}^{\texttt{m}}$ and confidence interval \par
$\mathcal{C}_{\alpha}^{\texttt{m}} = \left(\hat{\theta}_{\hat{\lambda},j}^{\texttt{m}} \pm z_{1-\alpha/2}\sqrt{\hat{\Sigma
}_{jj}/n}\right)$ for coordinate $j$} 
\caption{\texttt{multiPPI++}: Prediction-powered inference for Multinomial Classification}\label{alg:ppi_multi}
\end{algorithm}

\subsection{\texttt{multiPPI++}} 
\label{appendix:multiPPI}

In $K$-class multinomial classification problem with outcomes $Y_i\in \{0,\dots, K-1\}$ and covariates $X_i\in \mathbbm{R}^d$, the parameter of interests $\theta$ is a $d(K-1)$-dimensional vector whose $k$-th block $\theta_k\in \mathbbm{R}^d$ of components represent parameters for class $k \in \{1,\dots, K-1\}$ excluding the reference class to avoid overparameterization. The loss function in classical inference takes the form
\[l_{\theta}(X,Y) = -\frac{1}{n}\sum_{i=1}^n X_i^T\theta_{Y_i} + \log\left(\sum_{k=1}^{K-1} e^{X_i^T\theta_{k}}\right), \]

Defining \[\psi(\theta, x) = \log\left(\sum_{k=1}^{K-1} e^{x^T\theta_{k}}\right),\]
and the \texttt{multiPPI++} loss is given by
\[\begin{aligned}
L^{\texttt{m}}_{\lambda}(\theta) &= L_n(\theta) + 
\lambda (L_N^{f_u}(\theta)-L_n^{f_l}(\theta)) \\
&= - \frac{1}{n}\sum_{i=1}^n (X_{li}\theta_{Y_{li}}-\psi(\theta, X_{li}))\\
& -\frac{\lambda}{N}\sum_{i=1}^N(X_{ui}^T\theta_{\hat{Y}_{ui}^f} - \psi(\theta, X_{ui})) + \frac{\lambda}{n}\sum_{i=1}^n (X_{li}^T\theta_{\hat{Y}_{li}^f} - \psi(\theta, X_{li}))\\
\end{aligned}\]

\citet{angelopoulos2023ppi} shows that the parameter $\lambda$ can be optimally tuned to minimize the asymptotic variance of the prediction-powered estimate, leading to better estimation and inference than both the classical and \texttt{PPI} strategies. Particularly, in finite samples, we can obtain the plug-in estimate 

\begin{align*}
 \hat{\lambda} & = \frac{1}{2(1+\frac{n}{N})} \times \frac{\text{Tr} \left(\widehat{H}_{\hat{\theta}_{PPI}}^{-1}\left(\widehat{\text{Cov}}_{n}(\nabla l_{\hat{\theta}_{PPI}}, \nabla l_{\hat{\theta}_{PPI}}^f) + \widehat{\text{Cov}}_{n}( \nabla l_{\hat{\theta}_{PPI}}^f, \nabla l_{\hat{\theta}_{PPI}})\right)\widehat{H}_{\hat{\theta}_{PPI}}^{-1} \right)}{\text{Tr}\left(\widehat{H}_{\hat{\theta}_{PPI}}^{-1}\widehat{\text{Cov}}_{N+n}( \nabla l_{\hat{\theta}_{PPI}}^f)\widehat{H}_{\hat{\theta}_{PPI}}^{-1} \right)} 
\end{align*}

where $\hat{\theta}_{PPI} = \hat{\theta}_{\lambda}^{\texttt{PPI++}}$ is obtained by taking a fixed $\lambda \in [0,1]$. 

We summarize the \texttt{multiPPI++} adjusted inference procedures for multinomial logistic regression coefficients on predicted CODs by an NLP model $f$ against covariate $X$, under significance level $\alpha\in (0,1)$ in Algorithm~\ref{alg:ppi_multi}.

\newpage

\subsection{Parameter Estimates Across All Sites: Full Data 80/20 Split}

\label{sec:allsite_estimates}

\newcounter{pdfpage}
\setcounter{pdfpage}{1} % Start from page 1

% Loop to include images from the PDF
\loop
\begin{figure}[htbp]
 \centering
 \includegraphics[page=\thepdfpage]{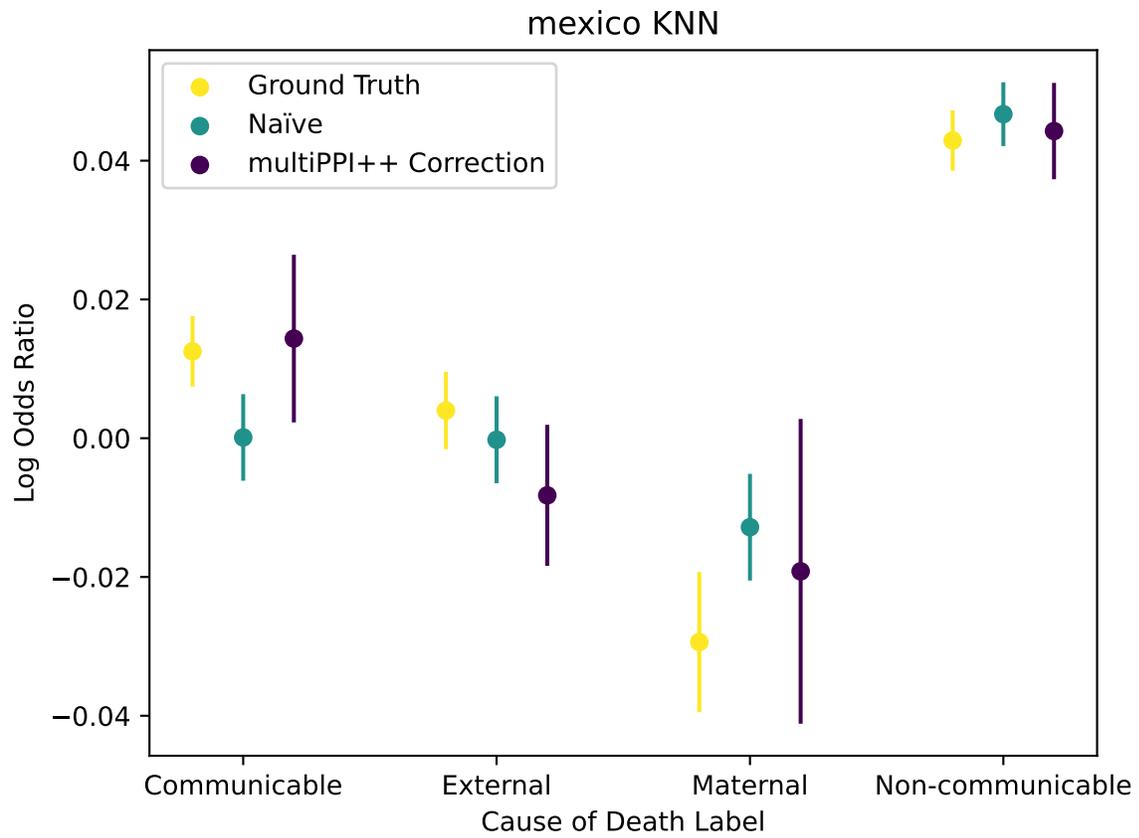}
 \caption{Full Data 80/20 Split: Site/Model \thepdfpage}
 \label{fig:page\thepdfpage1}
\end{figure}
\stepcounter{pdfpage} % Move to the next page
\ifnum\value{pdfpage}<22 % Change 22 to the total number of pages in your PDF
\repeat

\newpage

\subsection{Sensitivity Analysis}
In a sensitivity analysis, we varied our data splitting strategy, which revealed nuanced insights regarding the behavior of the classifier. Notably, when maintaining the split proportions (80/20) within each COD, the \texttt{PPI++} classifier exhibited minimal utilization of the labeled data, as indicated by the relatively small values of $\lambda$. Conversely, when employing an 80/20 split on the entire dataset, this lead to a significant information loss for the minority classes, resulting in larger $\lambda$ values for certain splits. Since an (80/20) on the whole dataset more closely resembles what one would see in reality on a prospective study while an (80/20) split by COD is resembles a retrospective study, this illustrates the importance of purposefully choosing an appropriate data splitting strategy to match the desired analysis. 
\newpage

\subsection{Parameter Estimates Across All Sites: Stratified 80/20 Split}

\label{sec:allsite_estimates_stratified}

\newcounter{pdfpage2}
\setcounter{pdfpage2}{1} % Start from page 1

% Loop to include images from the PDF
\loop
\begin{figure}[htbp]
 \centering
 \includegraphics[page=\thepdfpage]{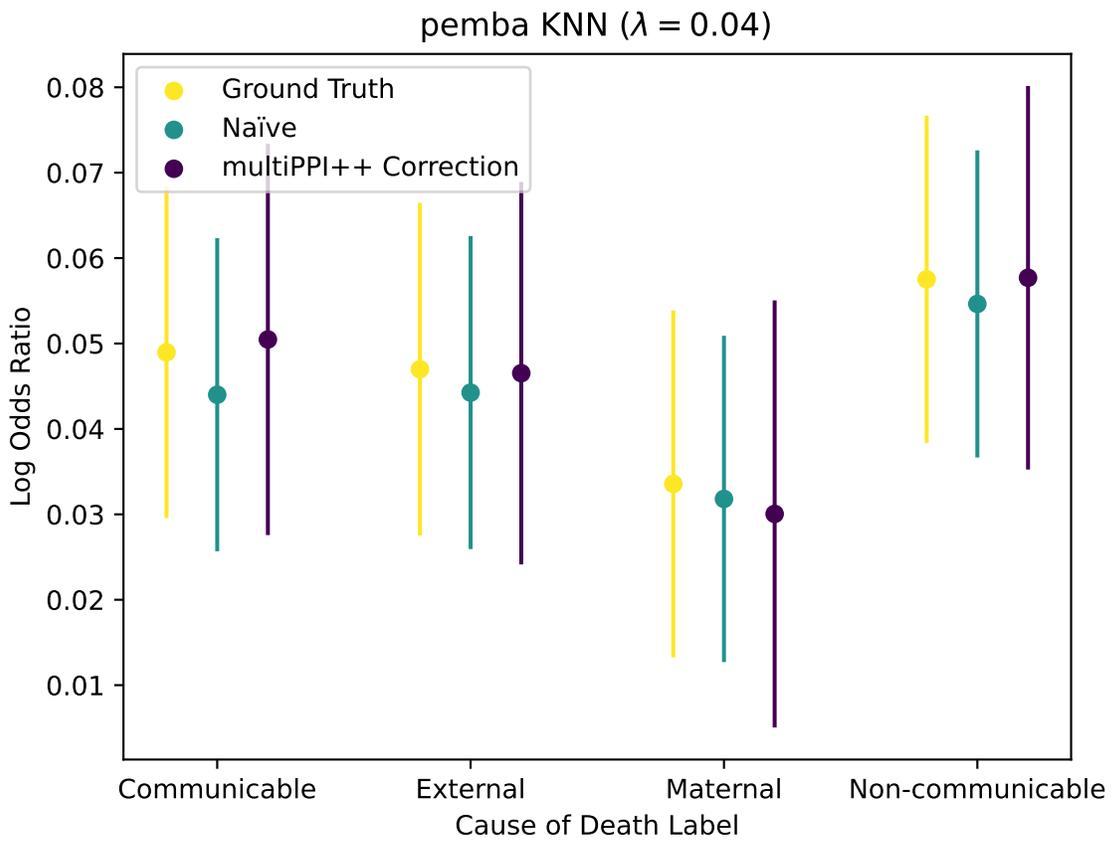}
 \caption{Stratified 80/20 Split: Site/Model \thepdfpage}
 \label{fig:page\thepdfpage2}
\end{figure}
\stepcounter{pdfpage2} % Move to the next page
\ifnum\value{pdfpage2}<27 % Change 22 to the total number of pages in your PDF
\repeat

\newpage
\section{Ethics Statement}
As this paper covers the study of the lives and deaths of real human beings, the ethical aspect of this investigation was considered of paramount importance. First, to ensure that any potential personally identifiable information was not mishandled, multiple layers of data security were employed, the primary means being the use of a secure Microsoft Azure cluster for the running of local models. When communication with outside organizations such as OpenAI's GPT models, we ensured that all data set were from the publicly available PHMRC data, which has been previously deidentified. Finally, in accordance with the importance of studying a sensitive topic such as VAs deserve, we have done our utmost to ensure that our study's objectives and tools utilized, to the best of our abilities, gives a fair, and respectful analysis of subject matter in question so that its results may be used for the furtherment of public health goals.

\section{Reproducibility Statement}
To ensure the reproducibility of our results, upon acceptance, we will publish our code repository with our entire pipeline - NLP training, predicting, inference and visualizations - to Github. In the meantime, during the review process, we enclose a zipped version of the code repository to Supplementary Materials through OpenReview. 

\end{document}